\documentclass[sigconf]{acmart}
\usepackage{verbatim} 
\usepackage[caption=false,font=footnotesize]{subfig}

\AtBeginDocument{%
  \providecommand\BibTeX{{%
    \normalfont B\kern-0.5em{\scshape i\kern-0.25em b}\kern-0.8em\TeX}}}

\copyrightyear{2023}
\acmYear{2023}
\setcopyright{acmlicensed}\acmConference[GECCO '23 Companion]{Genetic
and Evolutionary Computation Conference Companion}{July 15--19,
2023}{Lisbon, Portugal}
\acmBooktitle{Genetic and Evolutionary Computation Conference Companion
(GECCO '23 Companion), July 15--19, 2023, Lisbon, Portugal}
\acmPrice{15.00}
\acmDOI{10.1145/3583133.3596321}
\acmISBN{979-8-4007-0120-7/23/07}

\begin{document}

\title{Empirical Loss Landscape Analysis of Neural Network Activation Functions}

\author{Anna Sergeevna Bosman}
\affiliation{%
  \institution{Department of Computer Science, University of Pretoria}
  \streetaddress{Private bag X20}
  \city{Pretoria}
  \country{South Africa}
  \postcode{0028}
}
\email{anna.bosman@up.ac.za}
\orcid{0000-0003-3546-1467}

\author{Andries Engelbrecht}
\affiliation{%
  \institution{University of Stellenbosch}
  \city{Stellenbosch}
  \country{South Africa}
}
\affiliation{
\institution{Center for Applied Mathematics and Bioinformatics, Gulf University for Science and Technology}
\city{Kuwait City}
\country{Kuwait}
}
\email{engel@sun.ac.za}

\author{Mard{\'e} Helbig}
\affiliation{%
  \institution{School of Information and Communication Technology,\\ Griffith University}
  \streetaddress{G09 1.68, Gold Coast Campus}
  \city{Southport}
  \country{Australia}}
\email{m.helbig@griffith.edu.au}

\renewcommand{\shortauthors}{Bosman et al.}

\begin{abstract}
 Activation functions play a significant role in neural network design by enabling non-linearity. The choice of activation function was previously shown to influence the properties of the resulting loss landscape. Understanding the relationship between activation functions and loss landscape properties is important for neural architecture and training algorithm design. This study empirically investigates neural network loss landscapes associated with hyperbolic tangent, rectified linear unit, and exponential linear unit activation functions. Rectified linear unit is shown to yield the most convex loss landscape, and exponential linear unit is shown to yield the least flat loss landscape, and to exhibit superior generalisation performance. The presence of wide and narrow valleys in the loss landscape is established for all activation functions, and the narrow valleys are shown to correlate with saturated neurons and implicitly regularised network configurations.
\end{abstract}




\maketitle

\section{Introduction}
\label{intro}
The ability of artificial neural networks (NNs) to model non-linear functions of arbitrary complexity arises from the use of non-linear activation functions in the hidden layers. 
Kordos and Duch~\cite{ref:Kordos:2004} showed that the activation function used in a NN has a significant effect on the resulting loss landscape, but their study was limited to monotone (sigmoidal) and non-monotone bounded functions. Kordos and Duch~\cite{ref:Kordos:2004} concluded that non-monotone activation functions yield more complex loss landscapes than monotone activation functions, and that non-smooth monotone activation functions introduce more plateaus than smooth monotone activation functions.

Bounded activation functions such as sigmoid and hyperbolic tangent (TanH) are prone to saturation, which was shown to be detrimental to NN performance for shallow~\cite{ref:Rakitianskaia:2015} and deep~\cite{ref:Hochreiter:1998} architectures alike. Modern activation functions such as rectified linear unit (ReLU)~\cite{ref:Nair:2010} and exponential linear unit (ELU)~\cite{ref:Clevert:2016} are less prone to saturation, and thus became the primary choice for deep learning~\cite{ref:Arora:2018}. However, the effect of the activation function on the resulting NN loss landscapes is not well understood. Recent studies showed that on a bounded set in the weight space, the ReLU activation function yields strong piece-wise convexity around the isolated local minima on NN loss landscapes~\cite{ref:Milne:2018,ref:Rister:2017}. However, these theoretical studies rely on numerous limiting assumptions. To the best of the authors' knowledge, no dedicated studies of the loss landscapes associated with the ELU activation function exist to date.\looseness=-1

This paper investigates NN loss landscapes under various activation functions using the loss-gradient cloud visualisation technique~\cite{ref:Bosman:2019}. The following novel observations are made:
\begin{itemize}
	\item The presence of wide and narrow valleys in the loss landscape is established for all activation functions, and the narrow valleys are shown to be associated with saturated neurons and implicitly regularised NN configurations.
	\item Loss landscape modality, i.e. the total number of unique local minima, is not influenced by the choice of the activation function.
	\item ReLU is shown to exhibit the highest degree of convexity.
	\item ELU is shown to exhibit the least flatness, and to yield a loss landscape that is resilient to overfitting.
	
\end{itemize}

The rest of the paper is structured as follows: Section~\ref{sec:act:bg} provides the necessary background for the study. Section~\ref{sec:act:setup} discusses the experimental procedure. Section~\ref{sec:act:basins} presents the empirical analysis of the loss landscapes associated with various activation functions. Section~\ref{sec:act:conclusions} concludes the paper.
\section{Background}\label{sec:act:bg}
This study is concerned with the relationship between the activation functions employed in a NN architecture, and the resulting NN loss landscape properties. To that extent, the following topics are covered: Section~\ref{sec:bg:act} discusses the activation functions and their relevance. Section~\ref{sec:bg:loss} provides a literature review of the existing loss landscape studies in the context of activation functions. Section~\ref{sec:bg:fla} discusses the fitness landscape analysis techniques employed in this study.

\subsection{Activation functions}\label{sec:bg:act} 
An activation function is applied to the $net$ input signal received by a neuron in order to decide whether the neuron should fire or not. Non-linear activation functions introduce non-linearity to NNs, and thus enable the universal function approximation properties~\cite{ref:Hornik:1989}.

Activation functions fall into two categories: bounded and unbounded. One of the earliest smooth bounded activation functions is the sigmoid~\cite{ref:Rumelhart:1985}, defined as $f(net)= \frac{1}{1 + e^{-net}}$.
The sigmoid is linear around the origin, and saturates (approaches asymptotes) for inputs of large magnitude. Neuron saturation is generally undesirable, since the gradient is very weak near the asymptotes, and may cause stagnation in the training algorithms~\cite{ref:Rakitianskaia:2015}.


To alleviate saturation, the hyperbolic tangent function (TanH) can be used in place of the sigmoid, defined as $f(net)=\frac{e^{net}-e^{-net}}{e^{net}+e^{-net}}$.
TanH function, similarly to the sigmoid, saturates for large inputs, but its zero-centered range reduces saturation speed.

Bounded activation functions, due to their saturating gradients, are known to cause the vanishing gradient problem~\cite{ref:Hochreiter:1998}. The rectified linear activation function (ReLU)~\cite{ref:Nair:2010}, defined as $f(net)= \max(net, 0)$, is not bounded for positive inputs, and thus only saturates to the left of the origin.
This quality makes ReLU effective for deep learning~\cite{ref:Arora:2018}. 

ReLU has been criticised for the hard saturation (derivative of zero) for all negative inputs. While hard saturation is argued to be biologically plausible~\cite{ref:Glorot:2011}, the derivative of zero indicates that negative inputs will not yield any weight updates. Thus, if a particular neuron saturates (outputs zero for all inputs) during the course of training, the neuron will not be able to recover, and will not contribute anything to the final model.

Numerous adaptations of the ReLU activation have been proposed, most adding a slope with non-zero gradient for the negative inputs~\cite{ref:Nair:2010,ref:Xu:2015}. One of the more recent successful modifications of the ReLU is the exponential linear activation (ELU)~\cite{ref:Clevert:2016}, defined as
\begin{equation*}
f(net)= 
\begin{cases}
net& \text{if } net > 0,\\
e^{net} - 1 & \text{otherwise.}
\end{cases}
\end{equation*}\label{eq:nns:elu}
For the negative inputs, ELU uses an exponentially decaying curve. As a result, negative inputs have a non-zero derivative that gradually approaches the asymptote. Thus, negative inputs contribute to the final model, making NN training easier.

\subsection{Activation functions and loss landscapes}\label{sec:bg:loss} 
Liang et al.~\cite{ref:Liang:2018} showed theoretically that ReLU yields the presence of local optima with non-zero error in the loss landscape. In the same study, smooth strictly convex increasing activation functions, such as TanH and ELU, were shown to eliminate the local optima with non-zero loss. However, Liang et al.~\cite{ref:Liang:2018} did not support the conjectures with empirical evidence. Cao et al.~\cite{ref:Cao:2017} theoretically analysed 2-layer ReLU networks, and concluded that the resulting loss landscape contains flat regions around the critical points. Once again, no empirical evidence was provided. Laurent and Brecht~\cite{ref:Laurent:2018} studied the NN loss landscape associated with the hinge loss and ReLU activation functions, and found that the optima yielded by ReLU are generally non-differentiable. Sharp and flat minima were discovered, where sharp minima were always suboptimal. The limitation of Laurent and Brecht's study~\cite{ref:Laurent:2018} is that the hinge loss rather than the log likelihood loss typically used for NN training was analysed, and that no empirical evidence was provided. A more recent study by Liu~\cite{ref:Liu:2021} employed quadratic loss for the analysis of single-hidden-layer ReLU networks, and has shown theoretically and empirically that differentiable minima exist in ReLU networks, and that spurious local minima were not found for the MNIST and CIFAR-10 classification problems. Clearly, the effect of ReLU on the loss landscape is not yet completely understood. Specifically, the presence or absence of spurious local minima is a topic of much debate with conflicting results published~\cite{ref:Safran:2018}.

This study performs an empirical investigation of the loss landscapes associated with the various activation functions using a fitness landscape analysis approach, to investigate the soundness of some of the theoretical claims listed above. Specifically, loss-gradient clouds are used to visualise and analyse the minima present in the loss landscapes.

\subsection{Fitness landscape analysis}\label{sec:bg:fla} 
Any optimisation problem is associated with a corresponding \textit{fitness landscape}, i.e. the hypersurface formed by the objective function values of all the candidate solutions. Studying various properties of the fitness landscape, such as modality, ruggedness, neutrality, and searchability, is beneficial for optimisation problem understanding and optimisation algorithm design alike. Fitness landscape analysis (FLA) is a collection of techniques that aim to quantify fitness landscape properties of an optimisation problem~\cite{ref:Malan:2014, ref:Pitzer:2012}. FLA was originally proposed in the evolutionary context, and was thus applied to discrete binary search spaces. However, FLA has since been extended to continuous spaces~\cite{ref:Malan:2013b, ref:Munoz:2015, ref:Pitzer:2012, ref:Malan:2021}.
The rest of this section discusses the FLA concepts relevant to this study.

\subsubsection{Progressive gradient sampling}\label{subsec:fla:sample} 
By definition, a continuous solution space $X$ contains infinite solutions, therefore any feasible empirical investigation of $X$ has to rely on a finite sample $X'$. In the context of NNs, $X$ comprises all real-valued weight combinations. A random sample can be used to obtain $X'$. However, areas of high fitness (i.e. low error) are likely to not be adequately captured via random sampling, therefore important features of the landscape such as modality (properties of the minima) will not be represented~\cite{ref:Smith:2001, ref:Bosman:2018}.

An alternative to random sampling for NN loss landscapes is the progressive gradient walk (PGW)~\cite{ref:Bosman:2018,ref:Bosman:2019,ref:Bosman:2020}, used in this study. The PGW algorithm is based on the idea of a random walk~\cite{ref:Spitzer:2013, ref:Malan:2014b}. PGW is initialised at a random point in the search space. The sample is collected by performing consecutive steps through the solution space. Each successive step $\vec{x}_{l+1}$ is a combination of the current step $\vec{x}_{l}$ and some $\Delta\vec{x}_l$: $\vec{x}_{l+1} := \vec{x}_{l} + \Delta\vec{x}_l$. The direction of each step is determined by using the negative gradient of the loss function. For a step $\vec{x}_l$, the gradient vector $\vec{g}_l$ is calculated. Then, $\vec{g}_l$ is used to generate a binary mask $\vec{b}_l$:
	\begin{equation*}
	b_{lj} =\begin{cases}
	0 & \text{if $g_{lj}>0$},\\
	1 & \text{otherwise},
	\end{cases}
	\end{equation*}
	where $j \in \{1,\dots,m\}$ for $\vec{g}_l \in \mathbb{R}^m$. The next step $\vec{x}_{l+1}$ is obtained using the progressive random walk (PRW)~\cite{ref:Malan:2014b}. A single step of PRW constitutes randomly generating a step vector $\Delta\vec{x}_l\in \mathbb{R}^m$, such that $\Delta{x}_{lj}\in [0,\varepsilon]$~$\forall j \in
	\{1,\dots,m\}$, and determining the sign of each $\Delta{x}_{lj}$ using the corresponding ${b}_{lj}$ as
	\begin{equation*}
	\Delta{x}_{lj} :=\begin{cases}
	-\Delta{x}_{lj} & \text{if ${b}_{lj} = 0$},\\
	\Delta{x}_{lj} & \text{otherwise}.
	\end{cases}
	\end{equation*} 
 
To summarise, PGW randomises the magnitude of $\Delta\vec{x}_l$ per dimension, and sets the direction according to $\vec{g}_l$. Therefore, gradient information is combined with stochasticity, generating $X'$ that explores the areas of low error. 

\subsubsection{Loss-gradient clouds}\label{subsec:fla:lg} 
Loss-gradient clouds (LGCs) were first introduced in~\cite{ref:Bosman:2019} as a visualisation tool for the purpose of empirically establishing the presence and characteristics of optima in NN error landscapes. To construct LGCs, the weight space of a NN is sampled. Once an appropriate number of samples has been obtained, LGC is generated by constructing a scatter plot with loss (i.e. error) values on the $x$-axis, and gradient magnitude (i.e. norm) on the $y$-axis. Points of zero error and zero gradient correspond to global minima. Points of non-zero error and zero gradient correspond to local minima or saddle points, i.e. suboptimal critical points. The local convexity of the points can be further identified using Hessian matrix analysis: positive eigenvalues of the Hessian indicate convex local minima, and a mixture of positive and negative eigenvalues indicate saddle points. 

LGCs were successfully used to investigate the minima associated with different loss functions~\cite{ref:Bosman:2019} as well as NN architectures~\cite{ref:Bosman:2020}. This study is a natural extension of~\cite{ref:Bosman:2019, ref:Bosman:2020}, where the focus is shifted to the activation functions and their effect on the loss landscapes.

\section{Experimental Procedure}\label{sec:act:setup}
This study aimed to empirically investigate the modality, i.e. local minima and their properties, associated with different activation functions in the hidden layer of a NN. 
Feed-forward NNs with a single hidden layer and the cross-entropy loss function were used in the experiments. As previously discussed in Section~\ref{sec:bg:act}, TanH, ReLU, and ELU were considered for the hidden layer. 
For the output layer neurons, the sigmoid activation function was used. Section~\ref{sub:act:bench} below lists the benchmark problems used, and Section~\ref{sub:act:walks} outlines the sampling  parameters.

\subsection{Benchmark problems}\label{sub:act:bench}
Four real world classification problems of varied dimensionality and complexity were used in the experiments. Table~\ref{appendix:table:benchmarks} summarises the NN architecture parameters and corresponding NN dimensionality per dataset. Sources of each dataset and/or NN architectures are also specified. 
The following datasets were considered in this study:

\begin{enumerate}
\item{\bf XOR:} Exclusive-or (XOR) refers to the XOR logic gate, modelled by a NN with two hidden neurons. The dataset consists of 4 binary patterns only, but is not linearly separable.
\item{\bf Iris:} The Iris data set \cite{ref:Fisher:1936} is a collection of 50 samples per three species of irises: {\it Setosa}, {\it Versicolor}, and {\it Virginica}, comprising 150 patterns. 
\item{\bf Heart:} This is a binary classification problem  comprised of  920 samples, where each sample is a collection of various patient readings pertaining to heart disease prediction~\cite{ref:Prechelt:1994}.
\item{\bf MNIST:}
The MNIST dataset~\cite{ref:LeCun:2010} comprises 70000 examples of handwritten digits from $0$ to $9$. Each digit is stored as a $ 28\times 28 $ grey scale image. The 2D input is flattened into a 1D vector for the purpose of this study.
\end{enumerate}

\begin{table}[!tb]
\renewcommand{\arraystretch}{1}
\setlength{\tabcolsep}{3pt}
\caption{Benchmark Problems and the NN Architectures}
\label{appendix:table:benchmarks}
\centering
\begin{tabular}{lccccc}
\toprule
{\bf Problem} & {\# Input} & {\# Hidden} & {\# Output} & {Dimensionality} & {Source}\\
\midrule
XOR & 2 & 2 & 1 & 9& \cite{ref:Hamey:1998} \\
Iris & 4 & 4 & 3 & 35 &\cite{ref:Fisher:1936}  \\
Heart & 32 & 10 & 1 & 341 & \cite{ref:Prechelt:1994}\\
MNIST & 784 & 10 & 10 & 7960 & \cite{ref:LeCun:2010}\\
\bottomrule
\end{tabular}
\end{table}

For all problems except XOR, the inputs were $z$-score standardised. Binary classification problems used binary output encoding, while multinomial problems were one-hot encoded. All code used is available at \url{https://github.com/annabosman/fla-in-tf}.
\subsection{Sampling parameters}\label{sub:act:walks}

PGW was used to sample the areas of low error. To ensure adequate solution space coverage, the number of independent PGWs was set to $10\times m$, where $m$ is the dimensionality of the problem. Since gradient-based methods are sensitive to the starting point~\cite{ref:Bosman:2016}, two initialisation ranges for PGW were considered: $[-1,1]$ and $[-10,10]$. Malan and Engelbrecht~\cite{ref:Malan:2009} observed that the step size of the walk influences the resulting FLA metrics. As such, two step size settings were adopted in this study: maximum step size = 1\% of the initialisation range (micro), and maximum step size = 10\% of the initialisation range (macro). Micro walks were executed for 1000 steps, and macro walks were executed for 100 steps.

For all problems with the exception of XOR, the 80/20 training/testing split was used. The gradient ($G_t$) and the error ($E_t$) of the current PGW point was calculated using the training set. The generalisation error ($E_g$) was calculated on the test set for each point on the walk. For MNIST, a batch size of 100 was used. For the remainder of the problems, full batch (entire train/test subset) was used. To identify minima discovered by PGWs, the $G_t$ magnitude and $E_t$ value were recorded per step. Eigenvalues of the Hessian were calculated for all problems, except MNIST (due to computational constraints), to determine if a PGW point is convex, concave, saddle, or singular.

\section{Empirical Study of Modality}\label{sec:act:basins}
Empirical results of the study are presented in this section. Sections~\ref{sec:act:xor} to~\ref{sec:act:mnist} provide per-problem analysis of the three hidden neuron activation functions. 

\subsection{XOR}\label{sec:act:xor}

\begin{figure}[!tb]
	\begin{center}
		\begin{subfloat}[{TanH, micro, $[-1,1]$ initialisation. }\label{fig:xor:b1:tanh:micro}]{    \includegraphics[width=0.9\linewidth]{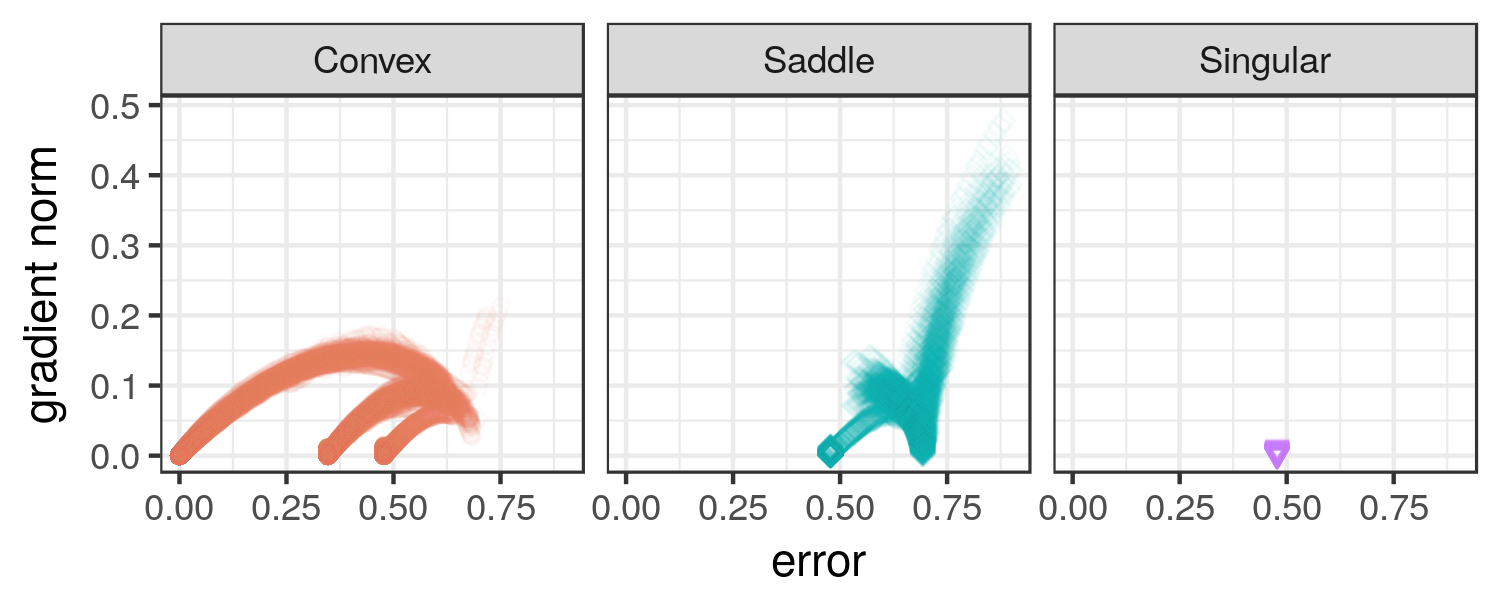}}
		\end{subfloat}
		\begin{subfloat}[{ReLU, micro, $[-1,1]$ initialisation. }\label{fig:xor:b1:relu:micro}]{    \includegraphics[width=0.9\linewidth]{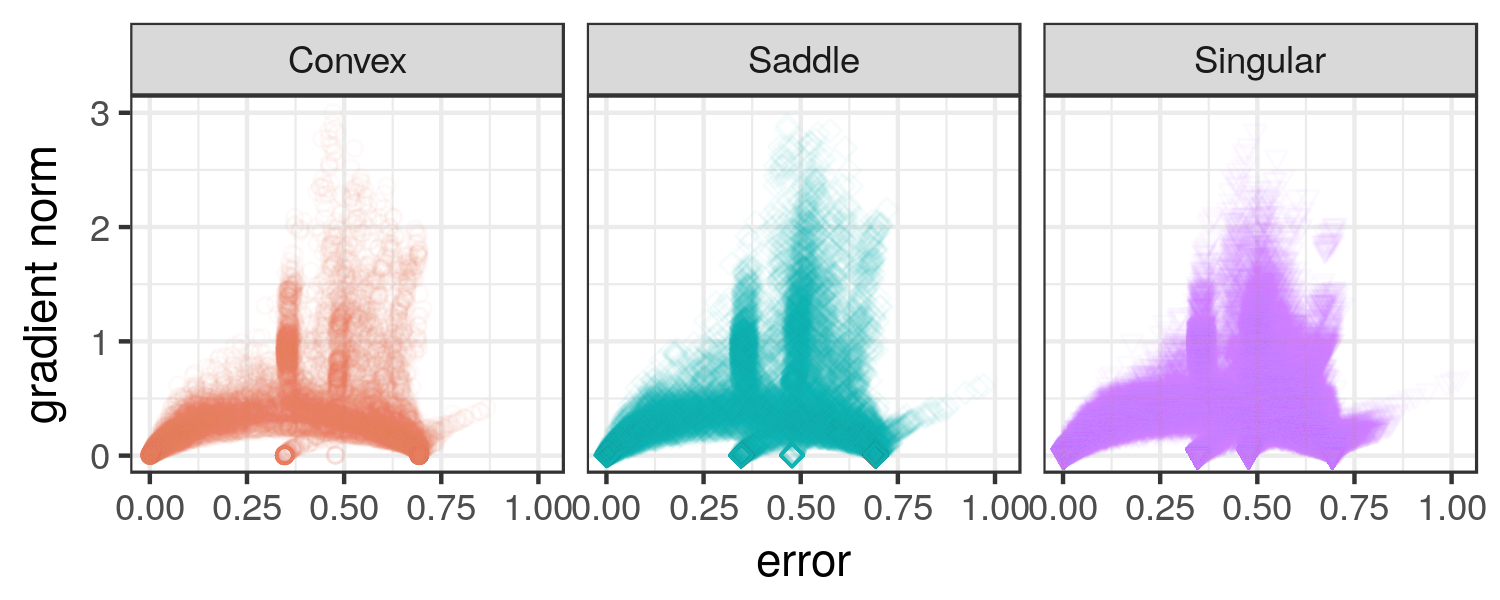}}
		\end{subfloat}
		\begin{subfloat}[{ELU, micro, $[-1,1]$ initialisation. }\label{fig:xor:b1:elu:micro}]{    \includegraphics[width=0.9\linewidth]{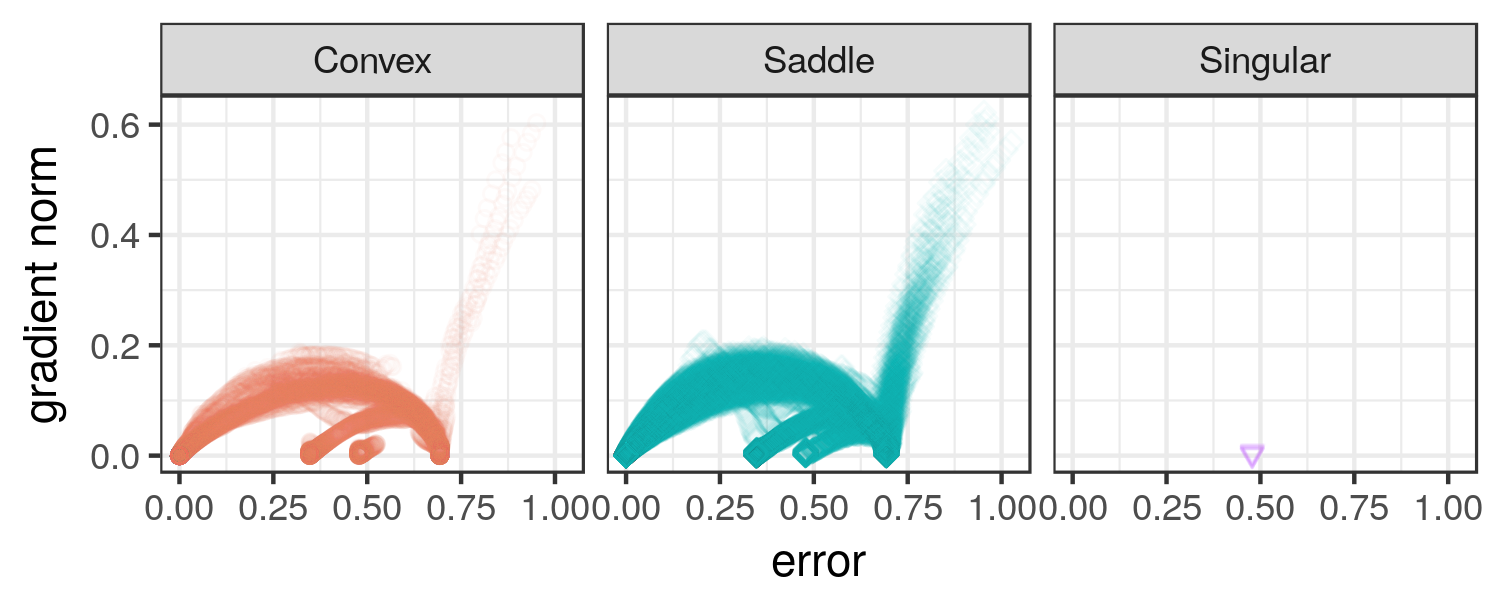}}
		\end{subfloat}
		\caption{LGCs for the micro PGWs for XOR.}\label{fig:xor:b1:micro:act}
	\end{center}
\end{figure}

\begin{figure}[!tb]
	\begin{center}
		\begin{subfloat}[{TanH, macro, $[-1,1]$ initialisation. }\label{fig:xor:b1:tanh:macro}]{    \includegraphics[width=0.9\linewidth]{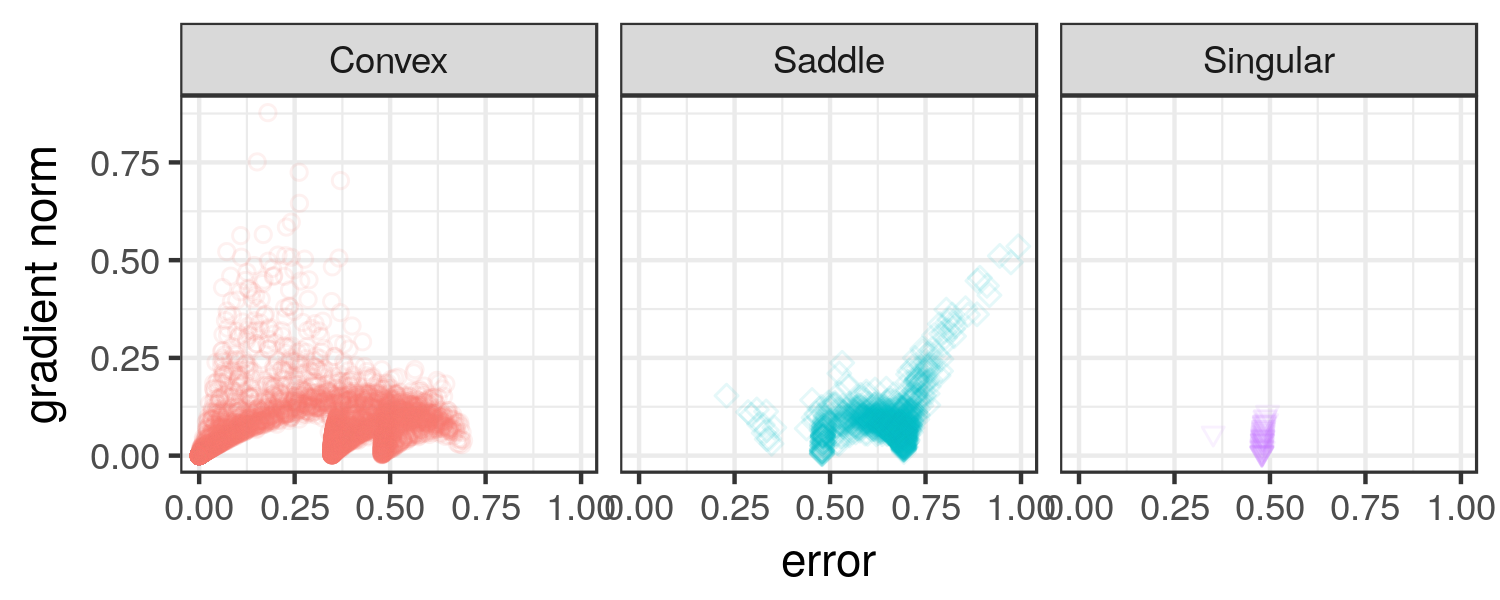}}
		\end{subfloat}
		\begin{subfloat}[{ReLU, macro, $[-1,1]$ initialisation. }\label{fig:xor:b1:relu:macro}]{    \includegraphics[width=0.9\linewidth]{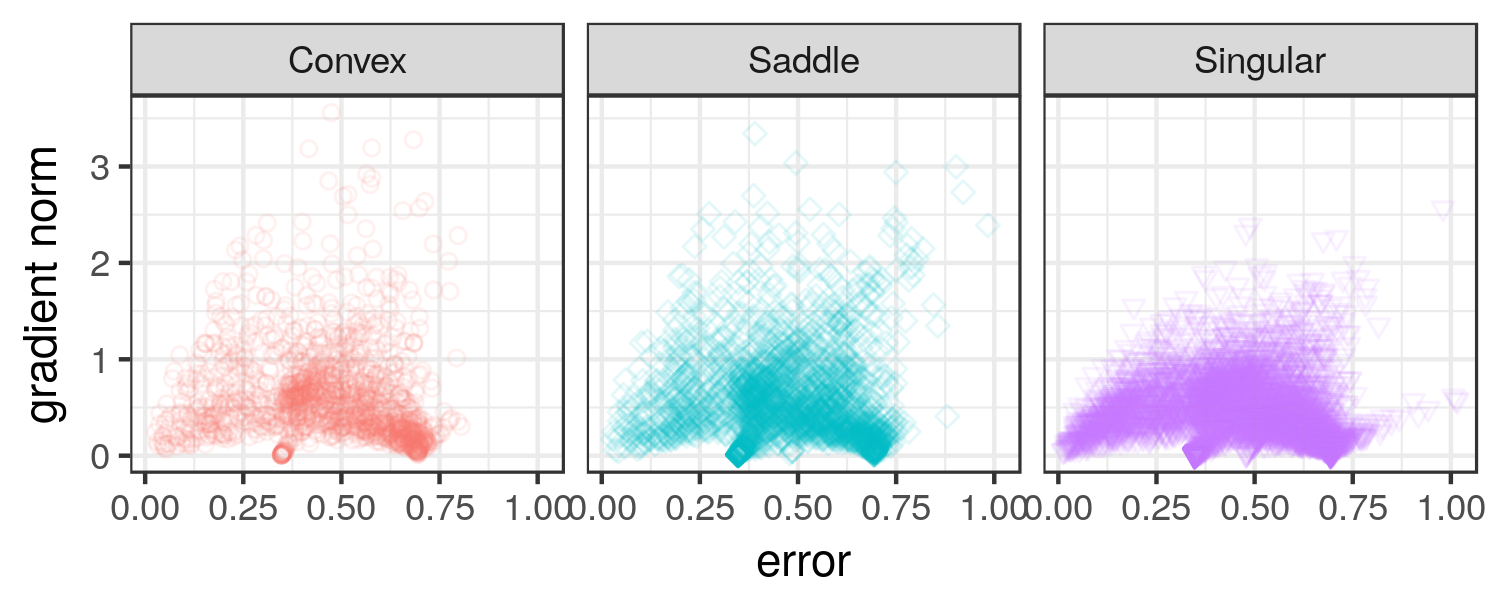}}
		\end{subfloat}
		\begin{subfloat}[{ELU, macro, $[-1,1]$ initialisation. }\label{fig:xor:b1:elu:macro}]{    \includegraphics[width=0.62\linewidth]{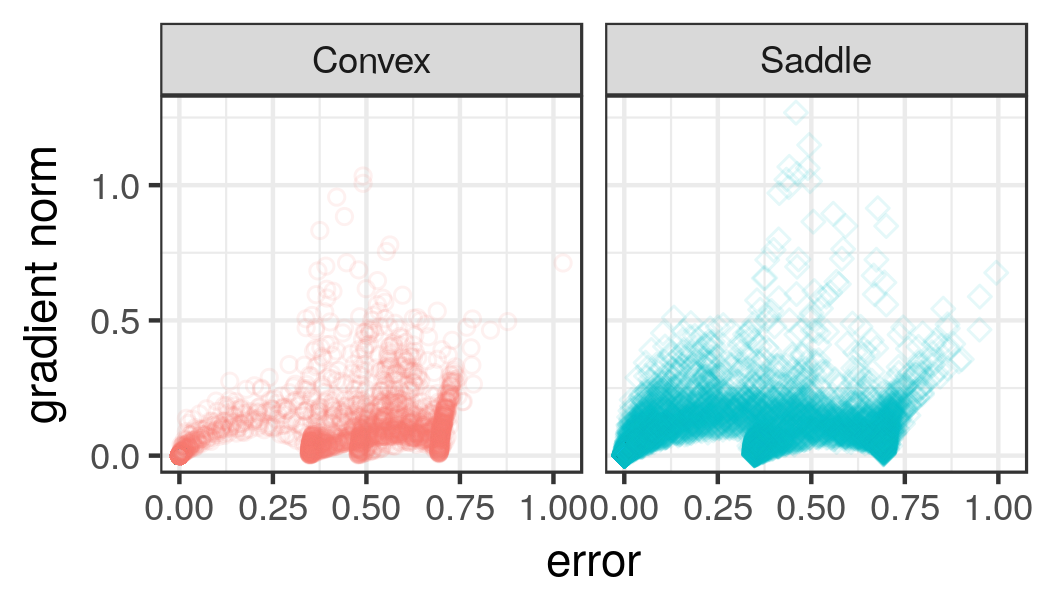}}
		\end{subfloat}
		\caption{LGCs for the macro PGWs for XOR.}\label{fig:xor:b1:macro:act}
	\end{center}
\end{figure}

Fig.~\ref{fig:xor:b1:micro:act} and \ref{fig:xor:b1:macro:act} show the LGCs for the XOR problem under the micro and macro settings. The sampled points are separated into panes according to curvature. Fig.~\ref{fig:xor:b1:micro:act} shows that the activation functions have all yielded four stationary attractors (points of zero gradient), three of which constituted local minima. However, the loss landscape characteristics around the local minima varied for the three activation functions. 


TanH exhibited a landscape with a clear transition from the saddle to the convex curvature, i.e. saddle and convex curvatures did not overlap. PGWs generally descended to a stationary point in the saddle space, and then made the transition to one of the convex minima. The three convex minima discovered were disconnected, i.e. the walks did not make transitions between the convex minima. 

For ReLU (Fig.~\ref{fig:xor:b1:relu:micro}) there was no clear separation between convex, saddle, and indefinite curvatures in the different areas sampled. Even though two convex local minima were discovered, these minima were evidently less attractive to PGWs than the global minimum. This observation indicates that a gradient-based algorithm is less likely to become trapped in a local minimum. The reason for this is probably the presence of saddle and indefinite (flat) points around the local minima, providing pathways through which a gradient-based algorithm can escape. Additionally, vertical clusters are observed around the local minima, indicating that diverse gradient information was available in the neighbourhood of the local minima. 
The predominance of indefinite curvature is explained by the fact that the ReLU function outputs zero for all negative inputs, which inevitably causes flatness.

ELU yielded a loss landscape similar to that of TanH, but with no clear separation between the convex and saddle curvatures. Four convex minima were discovered, but the suboptimal minima were again less attractive to PGWs than the global minimum. The local minima were less connected than for ReLU, but more connected than for TanH. 
The band connecting the global minimum to the stationary attractors of high error was wider for ELU than for TanH, indicating that the surface was smoother, and the connecting valleys were wider.

Fig.~\ref{fig:xor:b1:macro:act} shows the LGCs as sampled by the $[-1,1]$ macro walks. Larger step sizes did not prevent PGWs from discovering the same convex minima for TanH. Clear transition from the saddle to convex curvature is still observed in Fig.~\ref{fig:xor:b1:tanh:macro}. Stronger gradients were sampled around the global minimum attractor, indicating that larger steps bounced off the walls of the attraction basin. Larger steps also allowed PGWs to proceed directly to one of the minima, sometimes avoiding prior convergence to a saddle stationary point.

For ReLU, larger step sizes yielded very noisy behaviour, with less evident structure, as shown in Fig.~\ref{fig:xor:b1:relu:macro}. This indicates that the loss landscape associated with ReLU  was rugged and inconsistent when observed at a larger scale. Fewer walks have discovered convex global minima. Most points exhibited flatness. 
Thus, although ReLU offers a higher chance of escaping local minima, searchability of the loss landscape suffers, and becomes more dependent on the chosen step size. 

ELU (see Fig.~\ref{fig:xor:b1:elu:macro}) exhibited more evident structure than ReLU, and stronger gradients than TanH. Compared to TanH and ReLU, ELU did not exhibit any flatness, i.e. indefinite Hessians, which indicates that the loss landscape associated with ELU was more searchable. 

\begin{figure}[!b]
	\begin{center}
		\begin{subfloat}[{TanH, micro, $[-10,10]$ initialisation. }\label{fig:xor:b10:tanh:micro}]{    \includegraphics[width=0.9\linewidth]{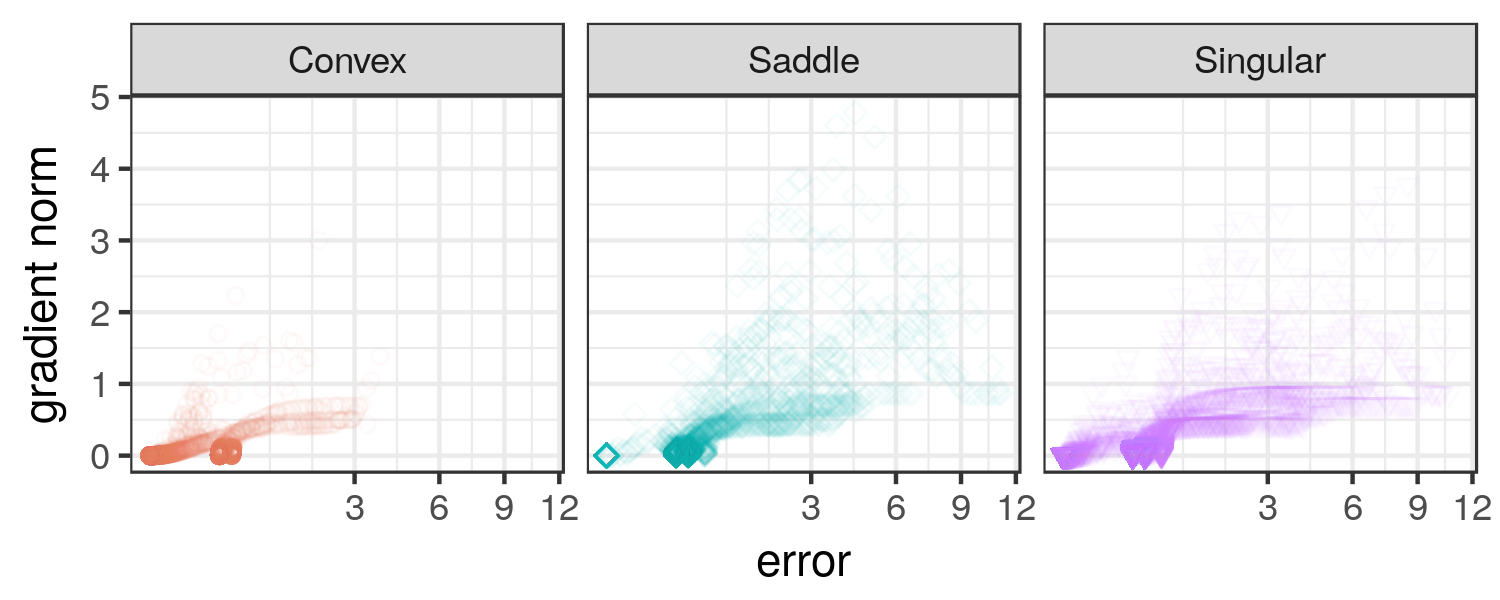}}
		\end{subfloat}
		\begin{subfloat}[{ReLU, micro, $[-10,10]$  initialisation.}\label{fig:xor:b10:relu:micro}]{    \includegraphics[width=0.9\linewidth]{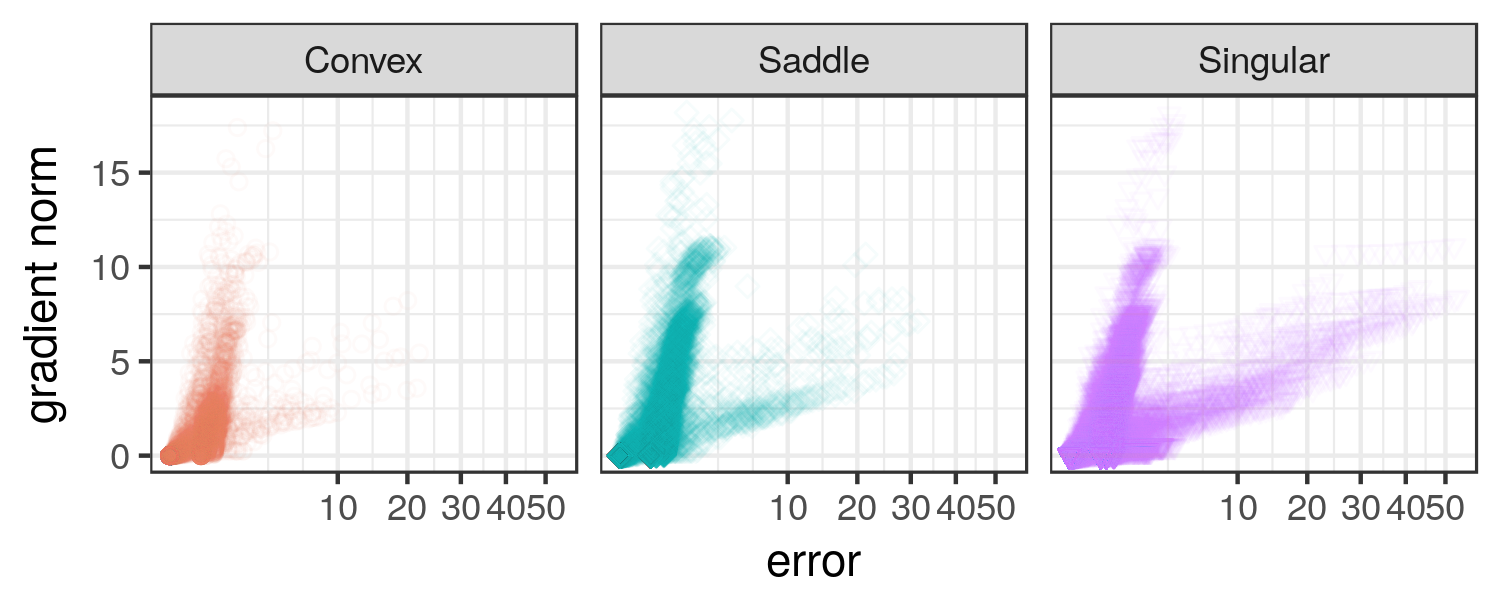}}
		\end{subfloat}
		\begin{subfloat}[{ELU, micro, $[-10,10]$  initialisation.}\label{fig:xor:b10:elu:micro}]{    \includegraphics[width=0.9\linewidth]{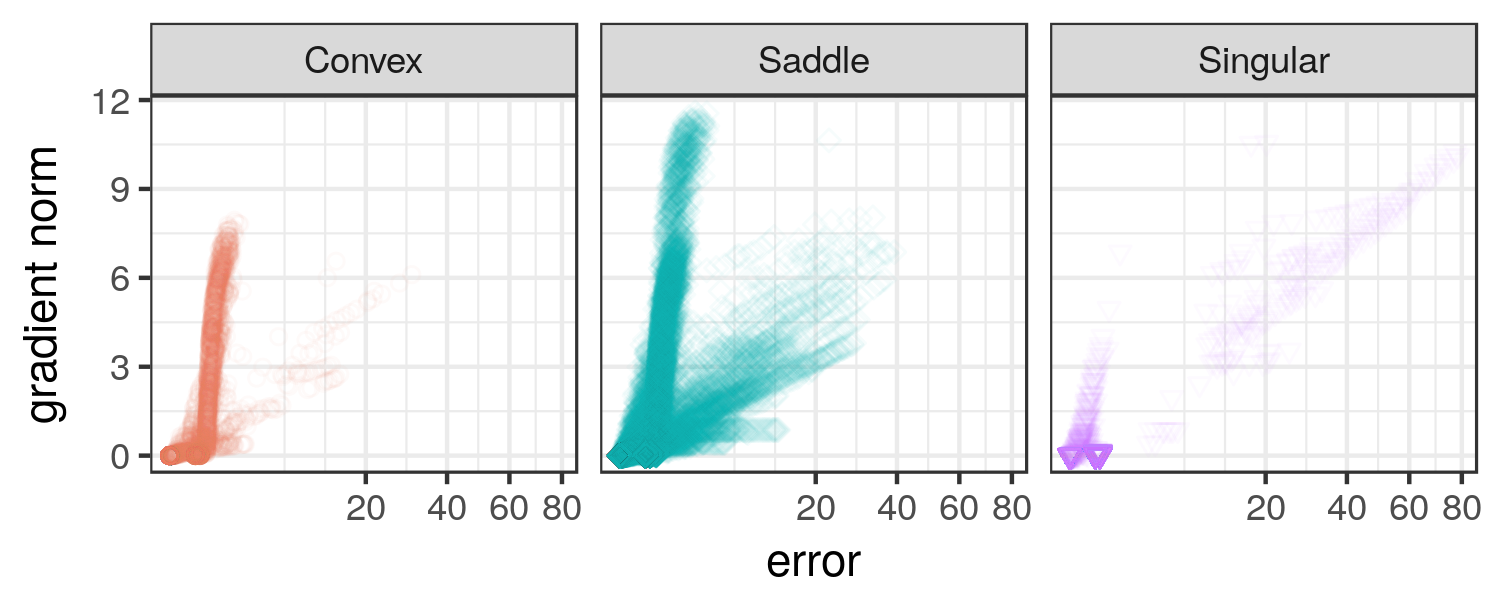}}
		\end{subfloat}
		\caption{LGCs for the micro PGWs for XOR.}\label{fig:xor:b10:micro:act}
	\end{center}
\end{figure}
Fig.~\ref{fig:xor:b10:micro:act} 
shows some of the LGCs obtained for the $[-10,10]$ PGWs. The $x$-axis is shown in square-root scale for readability. The same four stationary attractors were sampled for TanH, with two convex local minima. 
ReLU and ELU both yielded much higher errors than TanH, although convergence to the global minimum still took place. ReLU sampled the most flat curvature points, and ELU sampled the most saddle points. Both ReLU and ELU also exhibited two clusters: points of low error and high gradient, and points of low gradient and high error. These are attributed to the narrow and wide valleys present in the landscape, which could not be observed on the smaller scale. Out of the three activation functions, ELU exhibited the least number of indefinite curvature points. Thus, ELU yielded the most searchable loss landscape for the XOR problem.


\subsection{Iris}\label{sec:act:iris}

\begin{figure}[!b]
	\begin{center}
		\begin{subfloat}[{TanH, micro, $[-1,1]$  initialisation.}\label{fig:iris:b1:tanh:micro}]{    \includegraphics[width=0.9\linewidth]{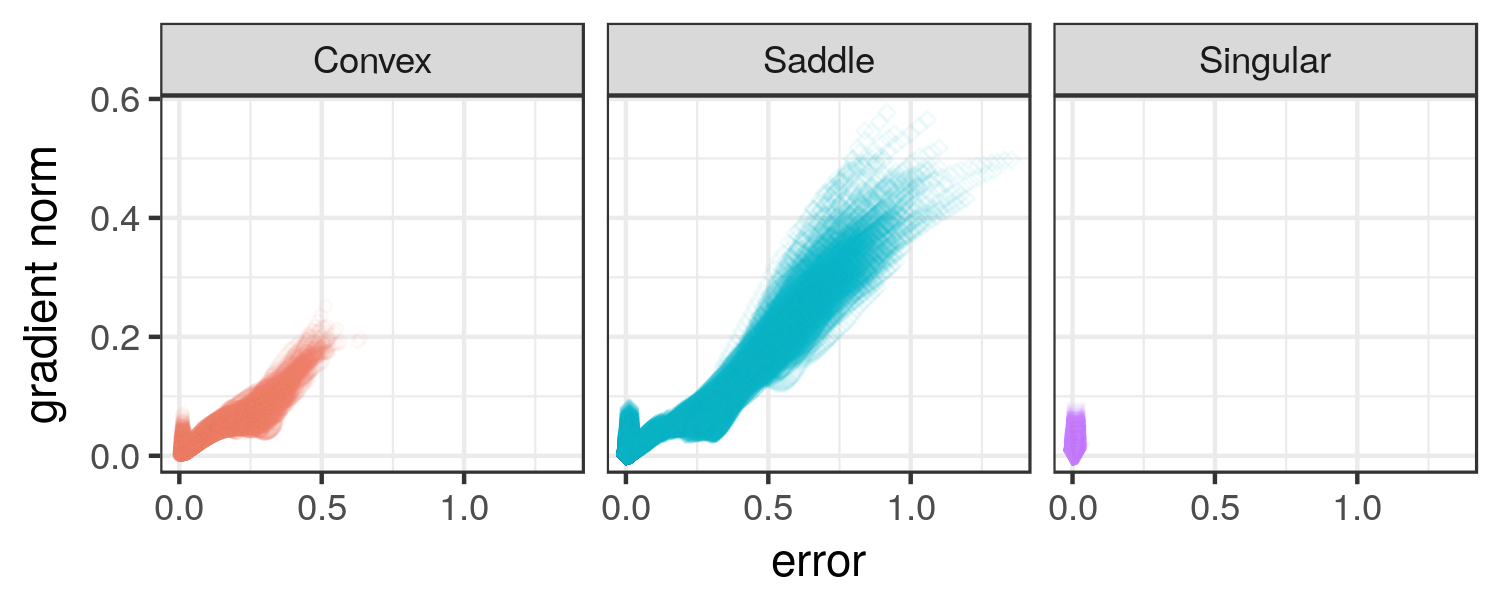}}
		\end{subfloat}
		\begin{subfloat}[{ReLU, micro, $[-1,1]$  initialisation.}\label{fig:iris:b1:relu:micro}]{    \includegraphics[width=0.9\linewidth]{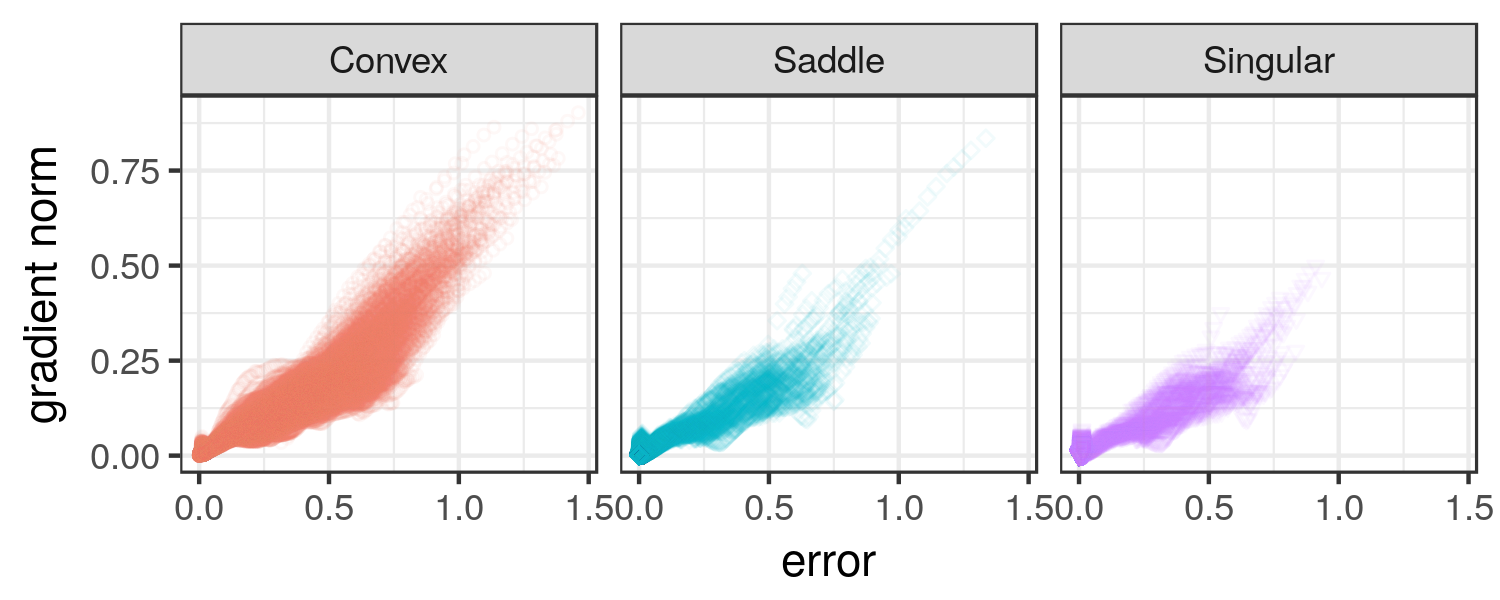}}
		\end{subfloat}
		\begin{subfloat}[{ELU, micro, $[-1,1]$  initialisation.}\label{fig:iris:b1:elu:micro}]{    \includegraphics[width=0.9\linewidth]{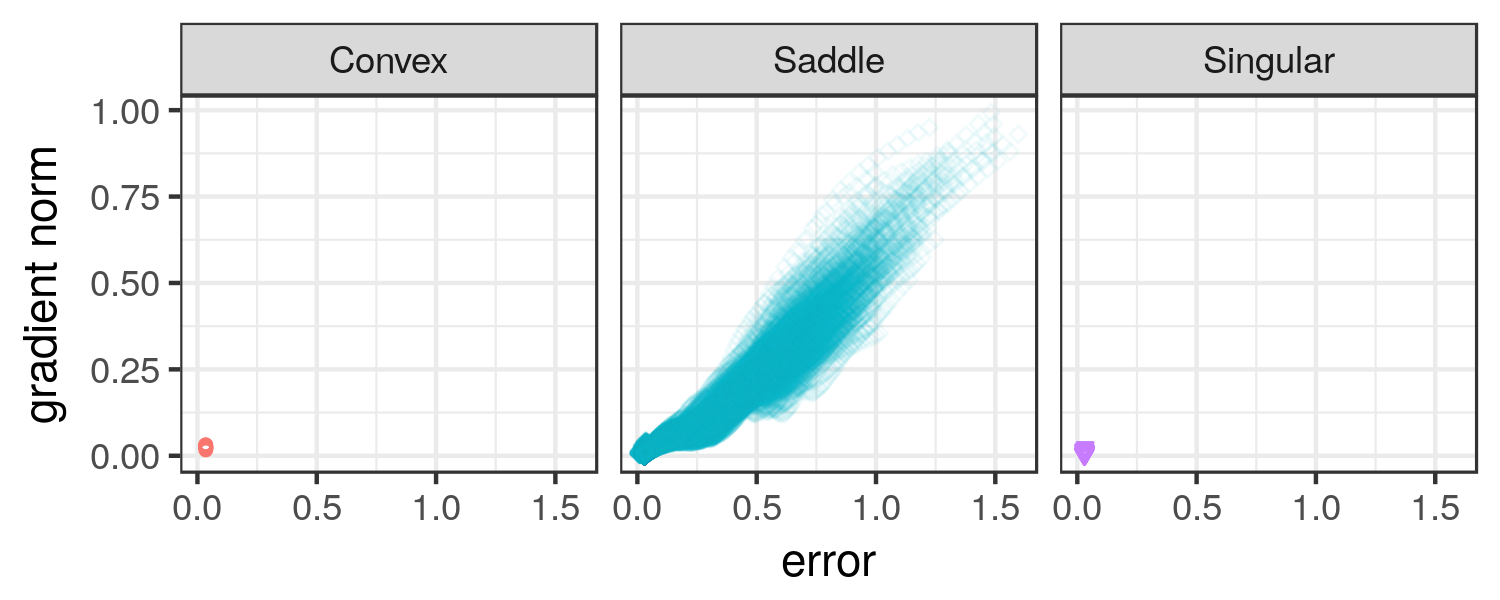}}
		\end{subfloat}
		\caption{LGCs for the micro PGWs for Iris.}\label{fig:iris:b1:micro:act}
	\end{center}
\end{figure}


Fig.~\ref{fig:iris:b1:micro:act} shows the LGCs obtained for the Iris problem sampled with the $[-1,1]$ micro walks. The LGCs for the three activation functions exhibited a similar shape, but different curvature properties. Only one major attractor around the global minimum was discovered. 

Fig.~\ref{fig:iris:b1:tanh:micro} shows that TanH was dominated by the saddle curvature. The sampled points were split into two clusters around the global minimum, namely points of higher error and lower gradient, and points of higher gradient and low error. The points of high gradient and low error overlapped with the points of indefinite curvature, indicating that this cluster exhibited flatness. Note that the flatness, i.e. lack of curvature in a particular dimension indicates that the corresponding weight did not contribute to the final prediction of the NN. For example, if a neuron is saturated, i.e. the neuron always outputs a value close to the asymptote, then the contribution of a single weight may become negligible. Therefore, indefinite curvature is likely associated with saturated neurons. However, neuron saturation is not the only explanation for non-contributing weights: if certain weights are unnecessary for the minimal solution to the problem at hand, then techniques such as regularisation can be employed to reduce the unnecessary weights to zero. Therefore, solutions with non-contributing weights can also be associated with regularised models. Smaller (regularised) architectures are ``embedded'' in the weight space of larger NN architectures, and can therefore be discovered by optimisation algorithms by setting the unnecessary weights to zero~\cite{ref:Mehta:2018}. A hypothesis is made that flat areas surrounding the global attraction basin correspond to the non-contributing weights, indicative of saturation. Some of the solutions with saturated neurons correspond to the minima that require fewer weights than available in the architecture. This hypothesis has the implication that the steep gradient attractor associated with indefinite curvature contains points of both poor and good performance, corresponding to unwanted saturation and implicit regularisation, respectively.

For ReLU, the prevalence of convex curvature is evident in Fig.~\ref{fig:iris:b1:relu:micro}. Again, ReLU exhibited more flatness than the other two activation functions. Such behaviour is attributed to the hard saturation of ReLU, which can easily yield non-contributing weights. ELU (see Fig.~\ref{fig:iris:b1:elu:micro}) was dominated by the saddle curvature. The least amount of flatness was discovered for ELU, making ELU a good choice for algorithms that rely on the gradient information. 

\begin{figure}[!b]
	\begin{center}
		\begin{subfloat}[{TanH, macro, $[-1,1]$  initialisation.}\label{fig:iris:b1:tanh:macro}]{    \includegraphics[width=0.9\linewidth]{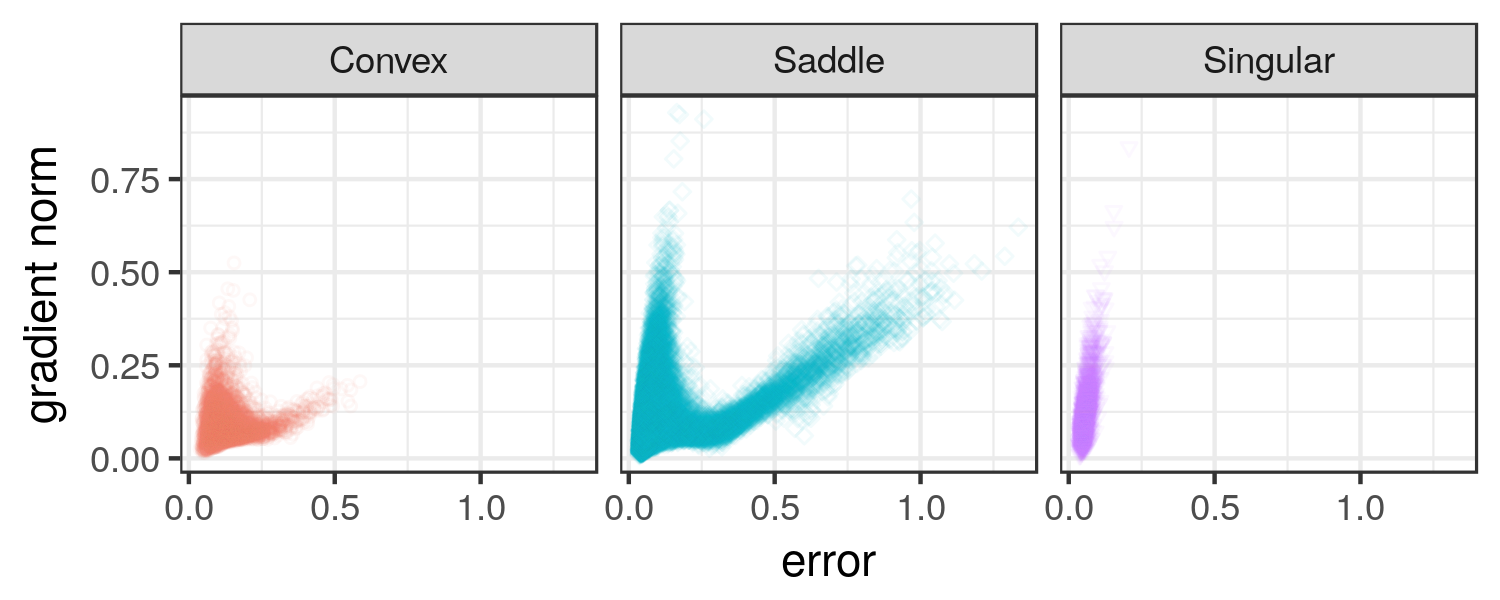}}
		\end{subfloat}
		\begin{subfloat}[{ReLU, macro, $[-1,1]$  initialisation.}\label{fig:iris:b1:relu:macro}]{    \includegraphics[width=0.9\linewidth]{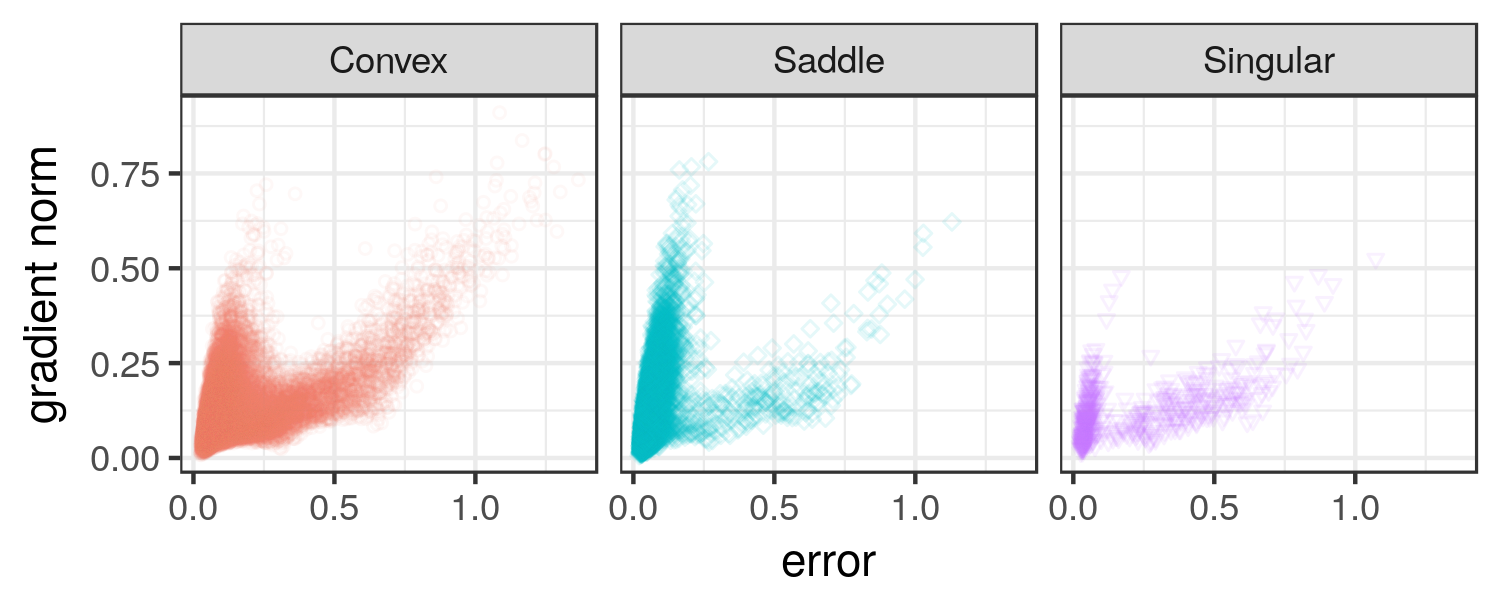}}
		\end{subfloat}
		\begin{subfloat}[{ELU, macro, $[-1,1]$  initialisation.}\label{fig:iris:b1:elu:macro}]{    \includegraphics[width=0.62\linewidth]{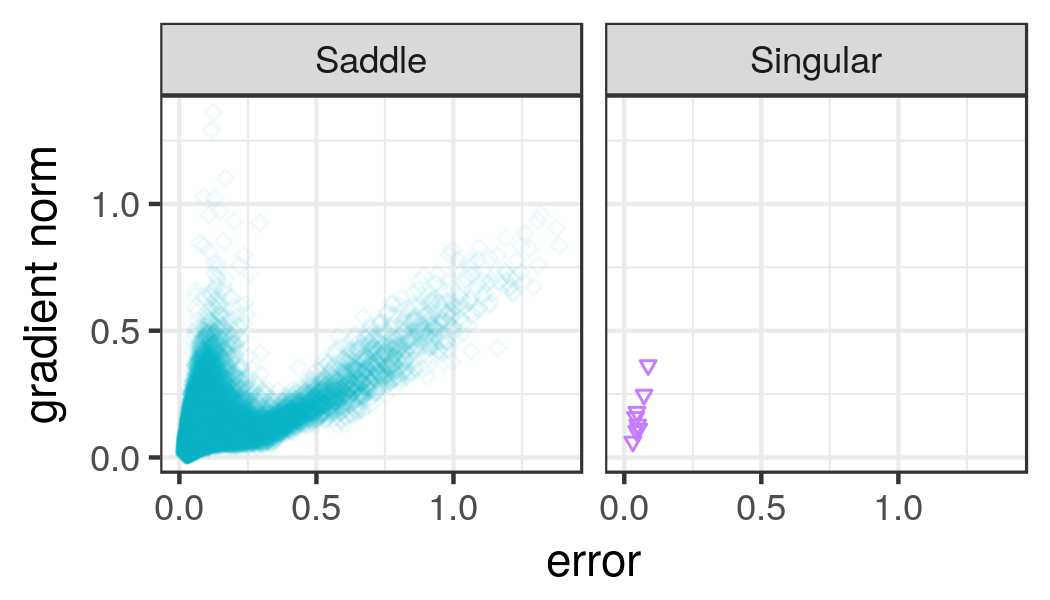}}
		\end{subfloat}
		\caption{LGCs for the macro PGWs for Iris.}\label{fig:iris:b1:macro:act}
	\end{center}
\end{figure}  

Fig.~\ref{fig:iris:b1:macro:act} shows the LGCs obtained for the $[-1,1]$ macro walks. Larger step sizes revealed similar curvature tendencies for the three activation functions, with TanH exhibiting convexity at the global minimum, but being dominated by saddle points otherwise, ReLU exhibiting strong convexity, and ELU exhibiting no convexity and almost no flatness. All activation functions yielded a split into high error, low gradient, and high gradient, low error clusters. The high gradient, low error clusters were associated with indefinite Hessians for the three activation functions. This behaviour is again attributed to the embedded minima that require fewer weights, as well as hidden unit saturation. Larger steps are likely to arrive at larger weights, thus increasing the chances of saturation.

\begin{figure}[!h]
	\begin{center}
		\begin{subfloat}[{TanH, macro, $[-1,1]$}]{    \includegraphics[width=0.35\linewidth]{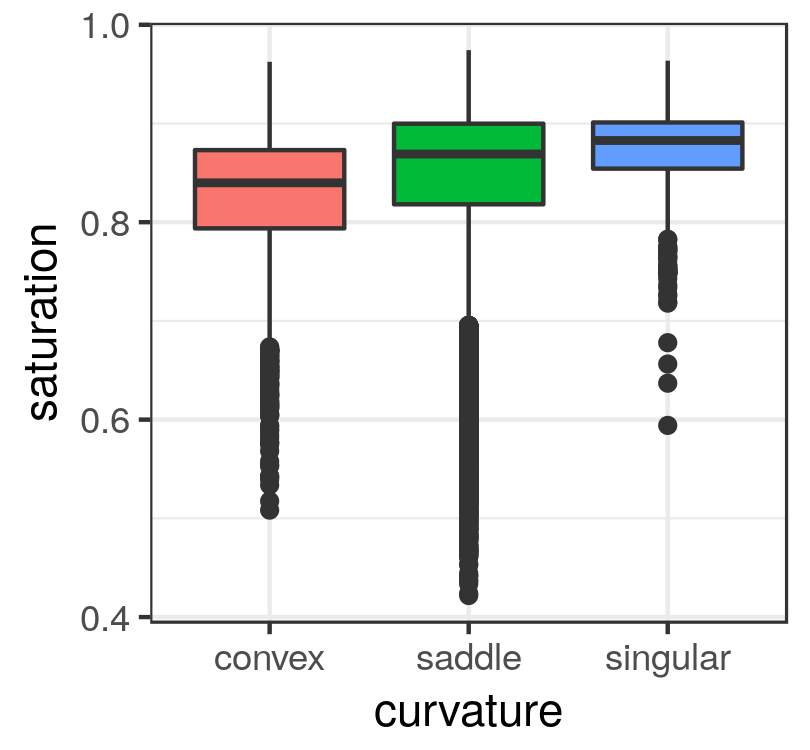}}
		\end{subfloat}
		\begin{subfloat}[{ReLU, macro, $[-1,1]$}]{    \includegraphics[width=0.35\linewidth]{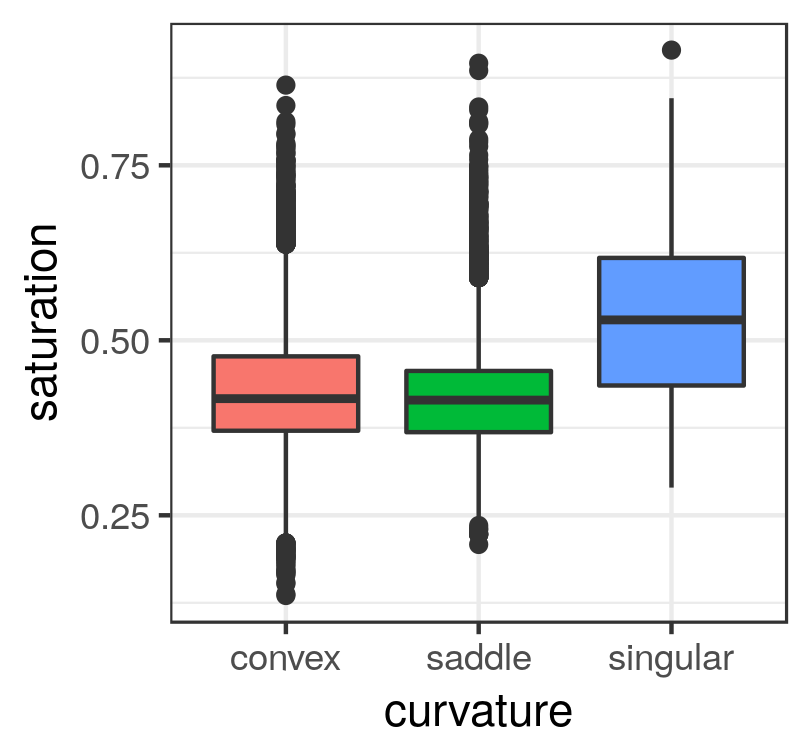}}
		\end{subfloat}\\
		\begin{subfloat}[{TanH, micro, $[-10,10]$}]{    \includegraphics[width=0.35\linewidth]{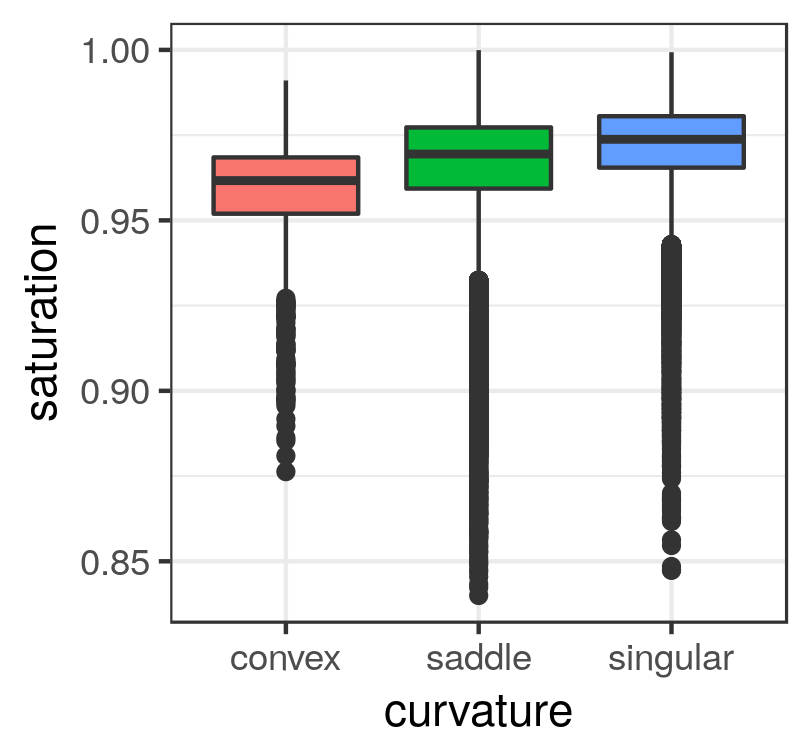}}
		\end{subfloat}
		\begin{subfloat}[{ReLU, micro, $[-10,10]$}]{    \includegraphics[width=0.35\linewidth]{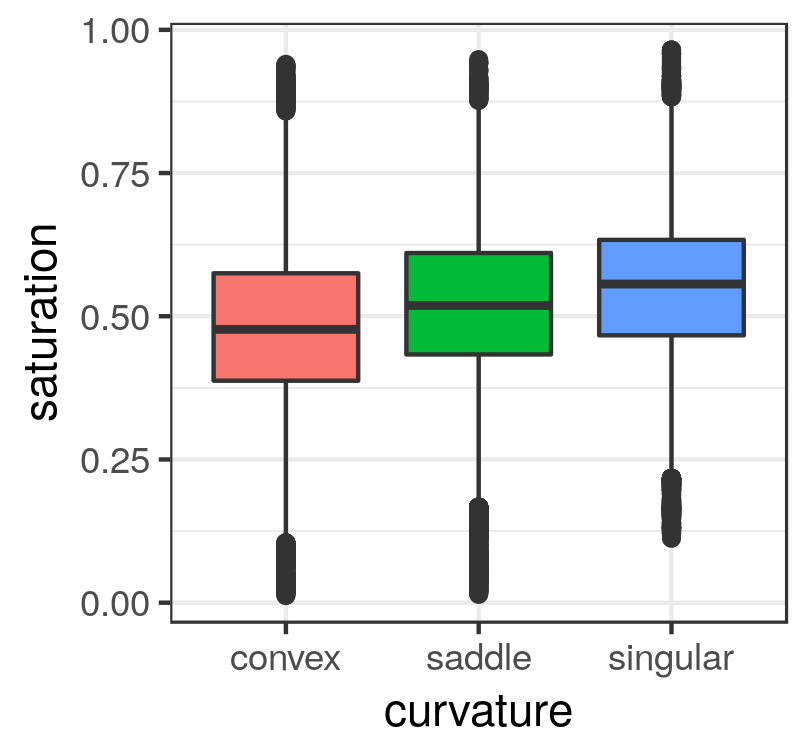}}
		\end{subfloat}
		\caption{Box plots illustrating the degree of saturation associated with the different curvatures for Iris.}\label{fig:iris:act:saturation}
	\end{center}
\end{figure}

To test the saturation hypothesis, the degree of saturation was measured for TanH and ReLU. For TanH, the $\varsigma_h$ saturation measure proposed in~\cite{ref:Rakitianskaia:2015b} for bounded activation functions was used. The value of $\varsigma_h$ is in the $[0,1]$ continuous range, where $0$ corresponds to a normal distribution of hidden neuron activations, $0.5$ corresponds to a uniform distribution of hidden neuron activations, and $1$ corresponds to a saturated distribution of hidden neuron activations, where most activations lie on the asymptotic ends. For ReLU, the proportion of zero activations for all hidden neurons was used as an estimate of saturation. Fig.~\ref{fig:iris:act:saturation} shows the box plots 
generated for TanH and ReLU for a few selected scenarios. 
Fig.~\ref{fig:iris:act:saturation} shows that singular curvature (indefinite Hessians) was indeed associated with higher saturation under various scenarios, especially under the $[-1,1]$ macro setting, which yielded a steep gradient cluster of indefinite points.

\begin{figure}[!tb]
	\begin{center}
		\begin{subfloat}[{TanH, macro, $[-10,10]$  initialisation.}\label{fig:iris:b10:tanh:macro}]{    \includegraphics[width=0.9\linewidth]{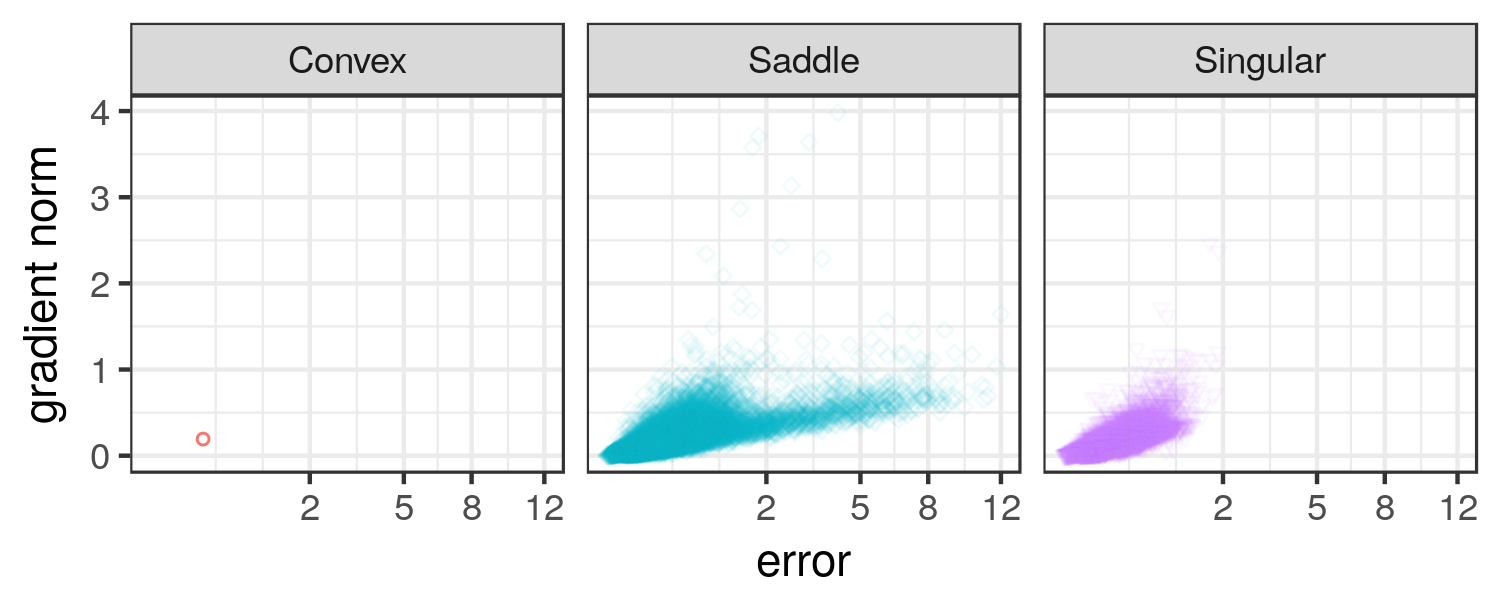}}
		\end{subfloat}
		\begin{subfloat}[{ReLU, macro, $[-10,10]$  initialisation.}\label{fig:iris:b10:relu:macro}]{    \includegraphics[width=0.9\linewidth]{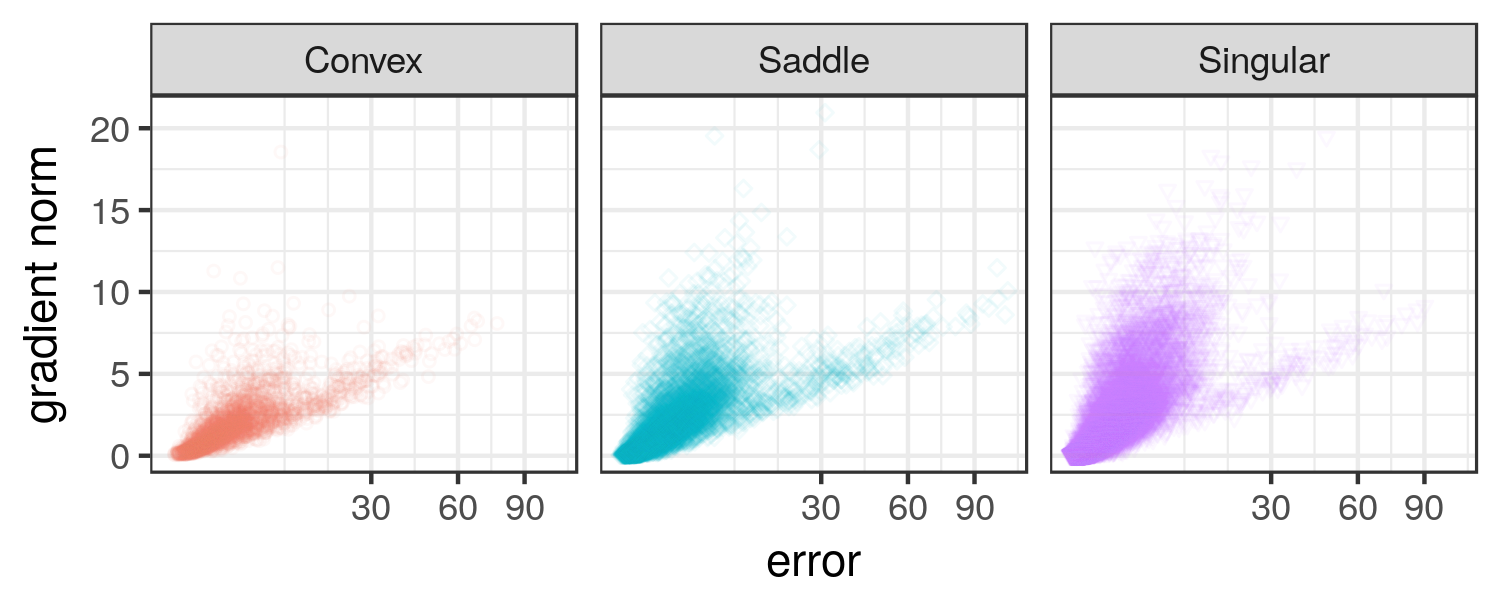}}
		\end{subfloat}
		\begin{subfloat}[{ELU, macro, $[-10,10]$  initialisation.}\label{fig:iris:b10:elu:macro}]{    \includegraphics[width=0.62\linewidth]{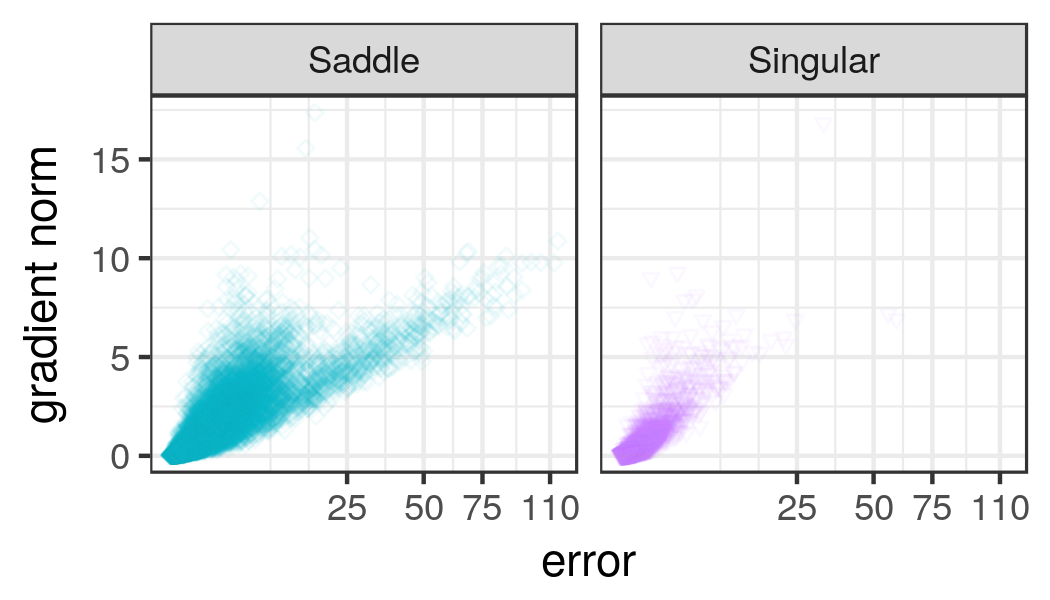}}
		\end{subfloat}
		\caption{LGCs for the macro PGWs for Iris.}\label{fig:iris:b10:macro:act}
	\end{center}
\end{figure}

Fig.~\ref{fig:iris:b10:macro:act} shows the LGCs for the $[-10,10]$ walks under the macro settings. The $x$-axis is shown in square-root scaling for readability. According to Fig.~\ref{fig:iris:b10:tanh:macro}, TanH exhibited very few convex points. Large step sizes prevented the walks from converging to a convex basin, indicating that the width of the convex basin must have been smaller than~$2$ (maximum  step size calculated as 10\% of the $[-10,10]$ initialisation range). ReLU exhibited the most convexity out of the three activation functions, and the most flatness. Thus, ReLU yielded a wider convex attraction basin than TanH, and was also more likely to saturate. ELU exhibited little to no convexity, and flatness only along the steeper cluster, associated with embedded minima and/or saturation. The split into two clusters was still evident for all activation functions. In general, with the increase in step size, the steep cluster became heavier than the shallow cluster. Thus, the steep cluster is confirmed to be associated with large weights, which cause saturation.

\begin{figure}[!tb]
	\begin{center}
		\begin{subfloat}[{TanH, $[-1,1]$}]{    \includegraphics[width=0.31\linewidth]{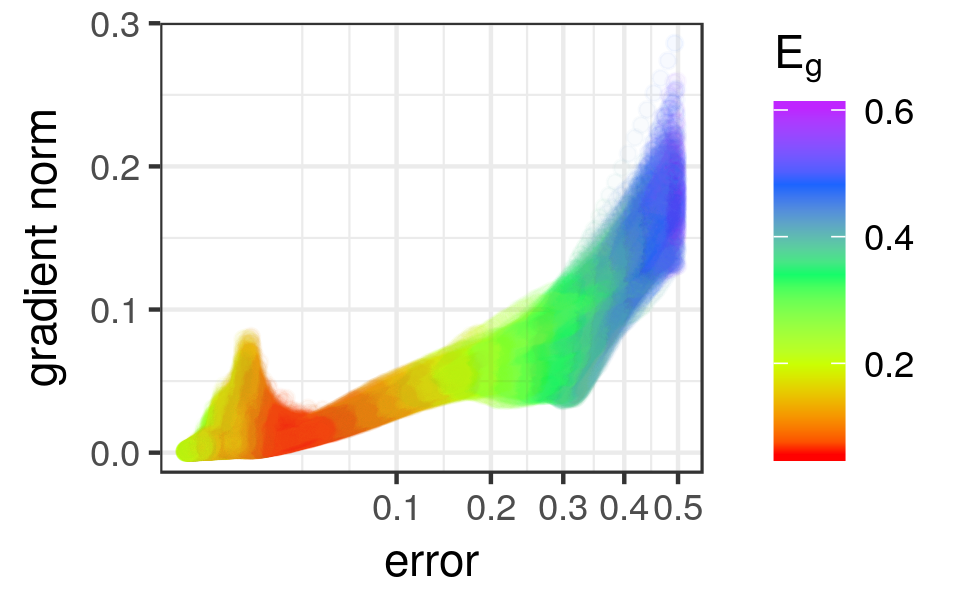}}
		\end{subfloat}
		\begin{subfloat}[{ReLU, $[-1,1]$}]{    \includegraphics[width=0.31\linewidth]{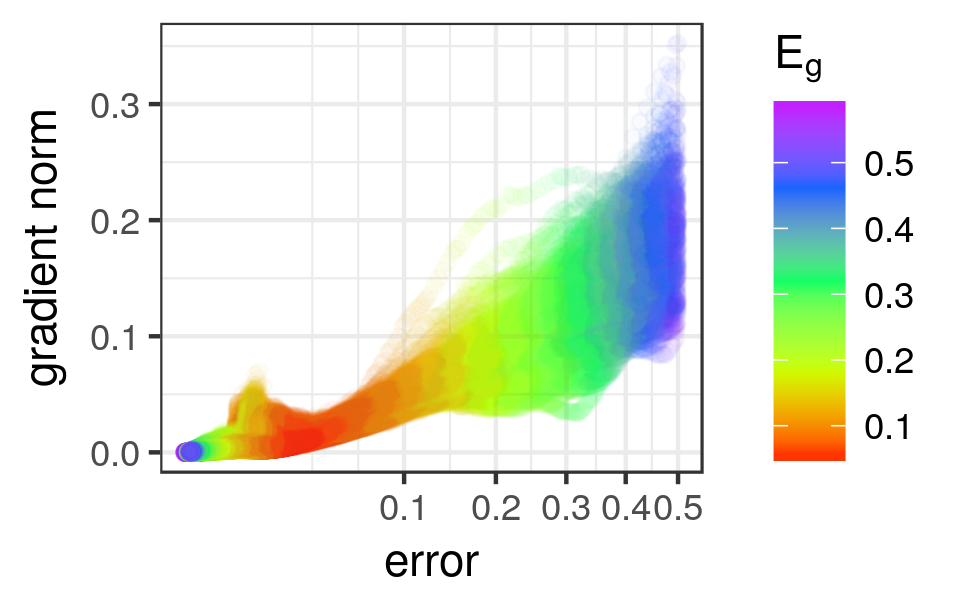}}
		\end{subfloat}
		\begin{subfloat}[{ELU, $[-1,1]$}]{    \includegraphics[width=0.31\linewidth]{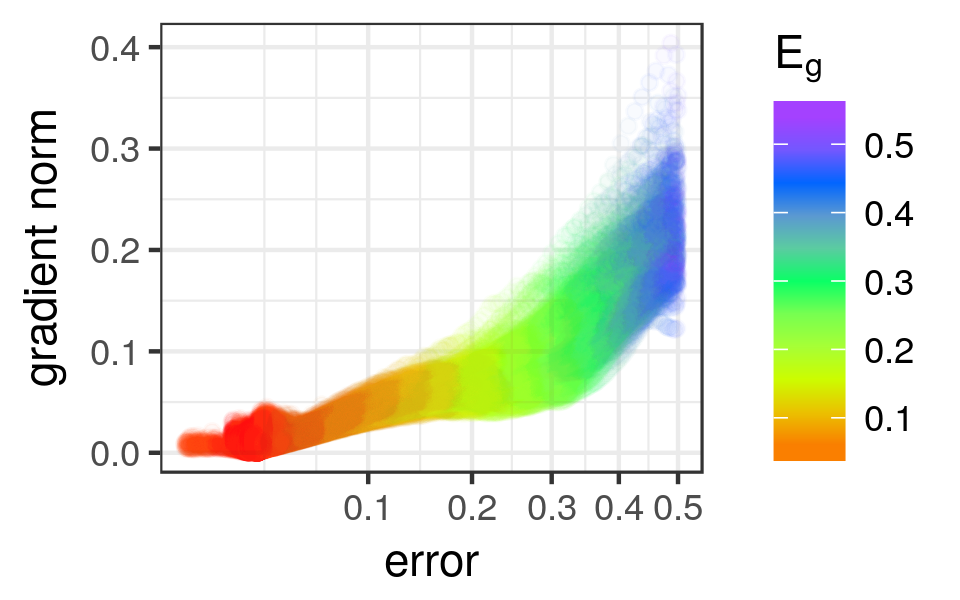}}
		\end{subfloat}
		\begin{subfloat}[{TanH, $[-10,10]$}]{    \includegraphics[width=0.31\linewidth]{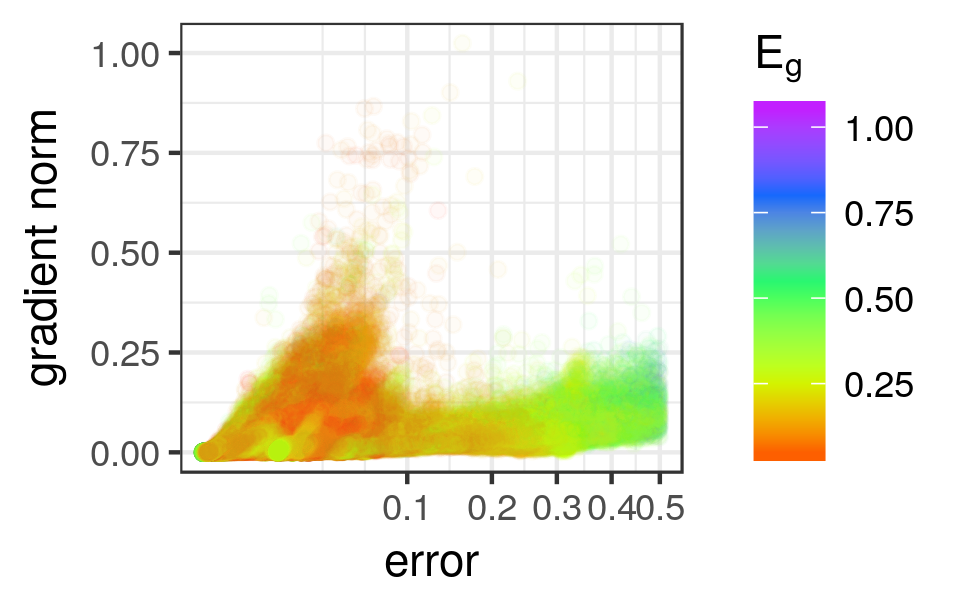}}
		\end{subfloat}
		\begin{subfloat}[{ReLU, $[-10,10]$}]{    \includegraphics[width=0.31\linewidth]{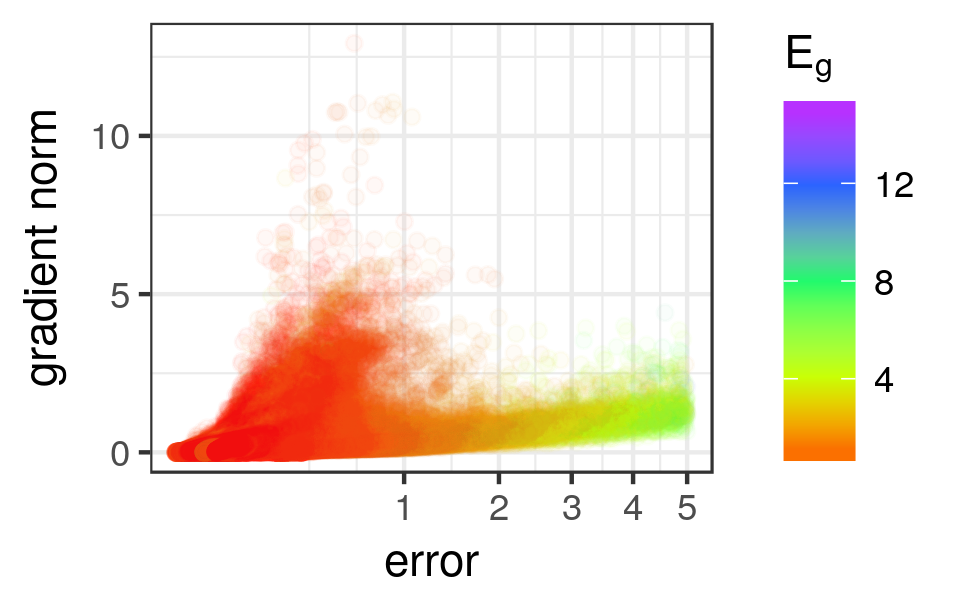}}
		\end{subfloat}
		\begin{subfloat}[{ELU, $[-10,10]$ }]{    \includegraphics[width=0.31\linewidth]{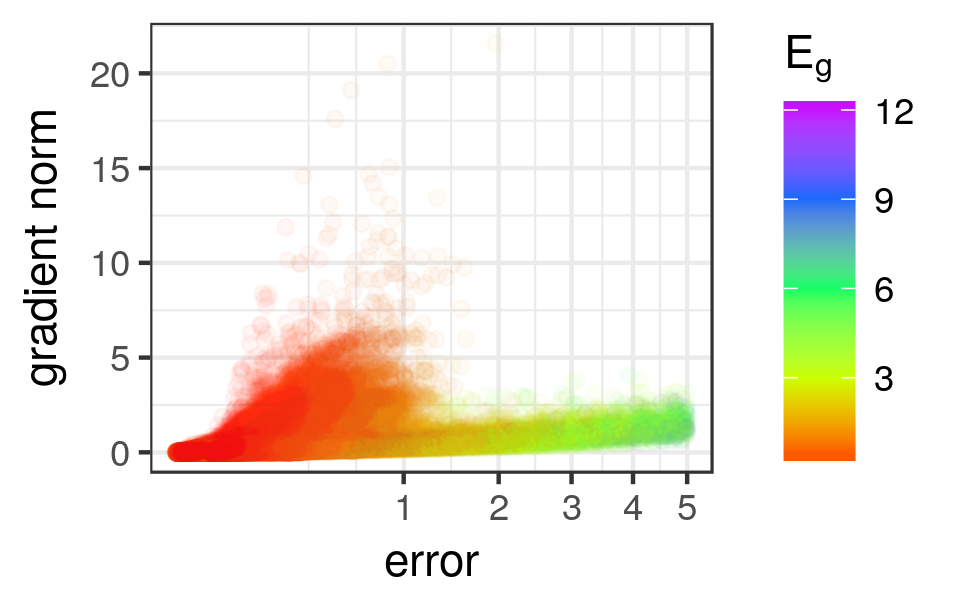}}
		\end{subfloat}
		\caption{LGCs colourised according to the $E_g$ values for Iris, micro PGWs, $E_t< 0.05$.}\label{fig:iris:act:gen}
	\end{center}
\end{figure}
To further study the generalisation behaviour of the three activation functions, Fig.~\ref{fig:iris:act:gen} shows the LGCs for the micro walks colourised according to their $E_g$ values. Fig.~\ref{fig:iris:act:gen} illustrates for the $[-1,1]$ setting that ELU exhibited the best generalisation behaviour. 
For the $[-10,10]$ setting, the low error, high gradient cluster generalised better than the high error, low gradient cluster for both ReLU and ELU. If the cluster of steep gradients corresponds to the embedded minima comprised of fewer contributing weights, then the steeper gradient solutions can be considered regularised, which explains better generalisation performance.
\subsection{Heart}\label{sec:act:heart}

\begin{figure}[!b]
	\begin{center}
		\begin{subfloat}[{TanH, macro}]{    \includegraphics[width=0.31\linewidth]{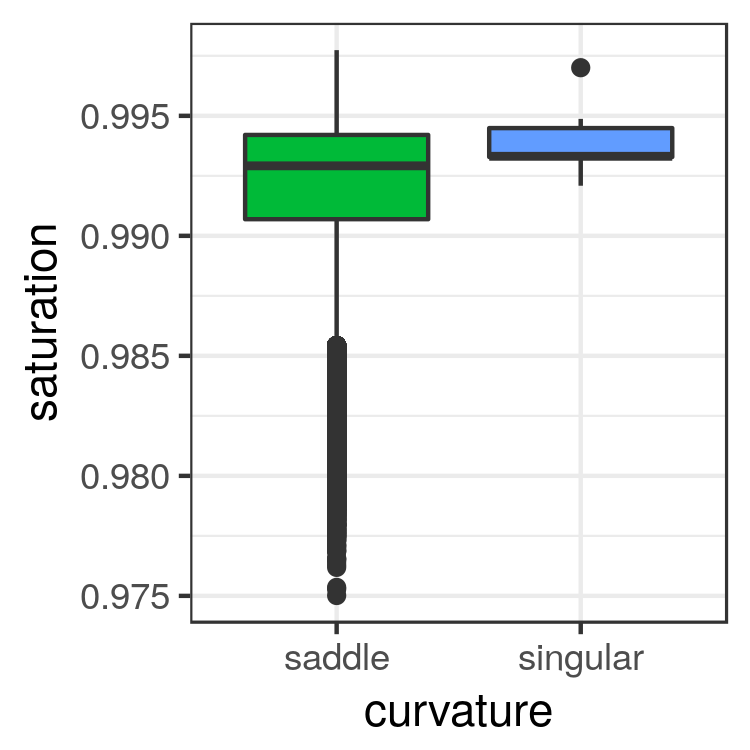}}
		\end{subfloat}	
		\begin{subfloat}[{ReLU, macro}]{    \includegraphics[width=0.31\linewidth]{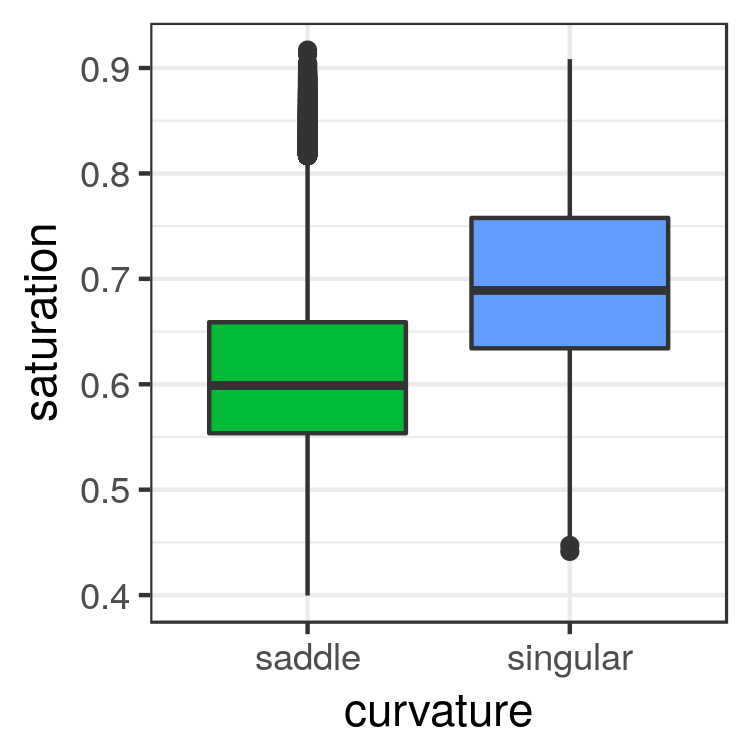}}
		\end{subfloat}
		\caption{Box plots illustrating the degree of saturation associated with the different curvatures for Heart, $[-10,10]$.}\label{fig:heart:act:saturation}
	\end{center}
\end{figure}

\begin{figure}[!b]
	\begin{center}
		\begin{subfloat}[{TanH, macro, $[-10,10]$}]{    \includegraphics[width=0.35\linewidth]{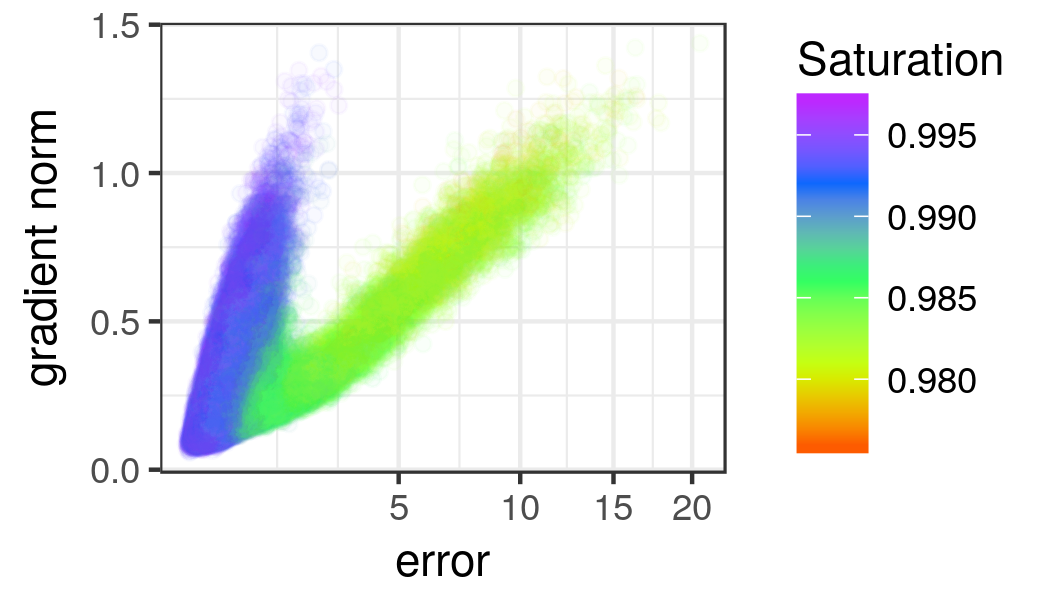}}
		\end{subfloat}	
		\begin{subfloat}[{ReLU, macro, $[-10,10]$}]{    \includegraphics[width=0.35\linewidth]{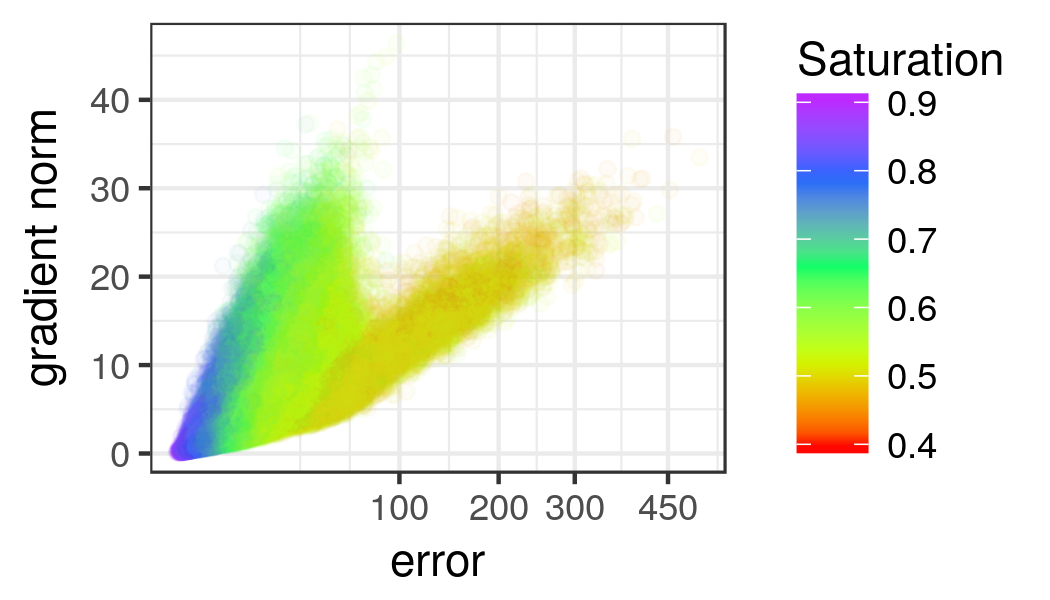}}
		\end{subfloat}
		\caption{LGCs for the macro PGWs initialised in the $[-10,10]$ range for Heart, colourised according to saturation.}\label{fig:heart:act:saturation:cloud}
	\end{center}
\end{figure}

For the heart problem, the split into two clusters was once again evident for all activation functions (figures not shown for brevity). 
Fig.~\ref{fig:heart:act:saturation} illustrates that the indefinite points corresponded to points of higher saturation for both TanH and ReLU. Further, Fig.~\ref{fig:heart:act:saturation:cloud} shows the LGCs for TanH and ReLU, colourised according to the estimated degree of saturation. For both activation functions, the steeper gradient cluster was clearly associated with the saturated neurons.

\begin{figure}[!tb]
	\begin{center}
		\begin{subfloat}[{TanH}]{    \includegraphics[width=0.31\linewidth]{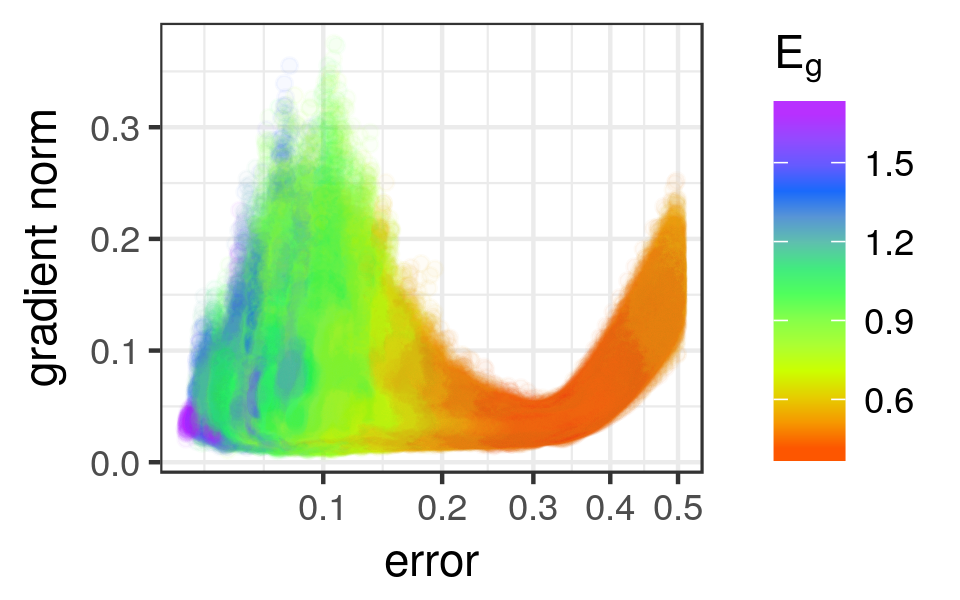}}
		\end{subfloat}
		\begin{subfloat}[{ReLU}]{    \includegraphics[width=0.31\linewidth]{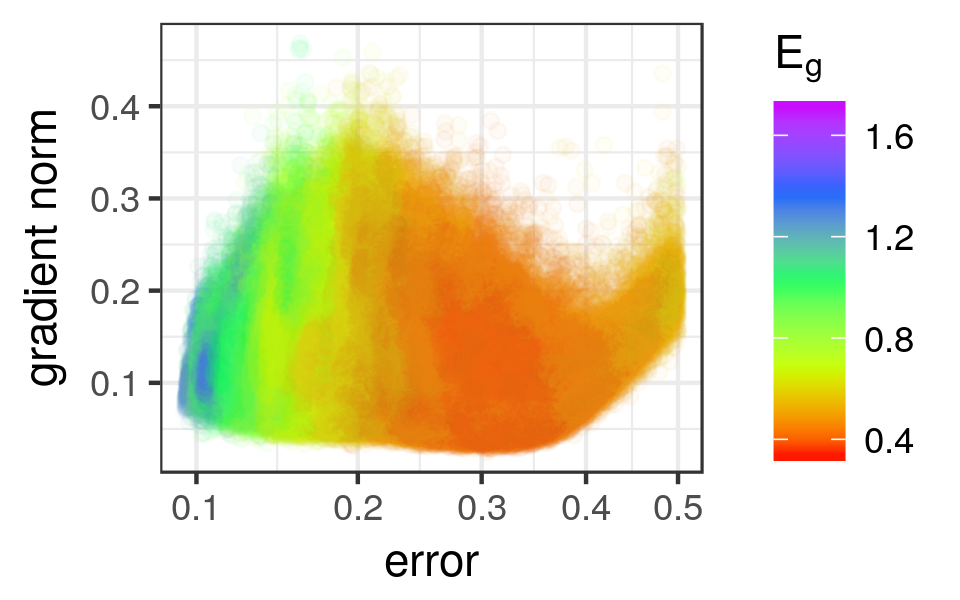}}
		\end{subfloat}
		\begin{subfloat}[{ELU}]{    \includegraphics[width=0.31\linewidth]{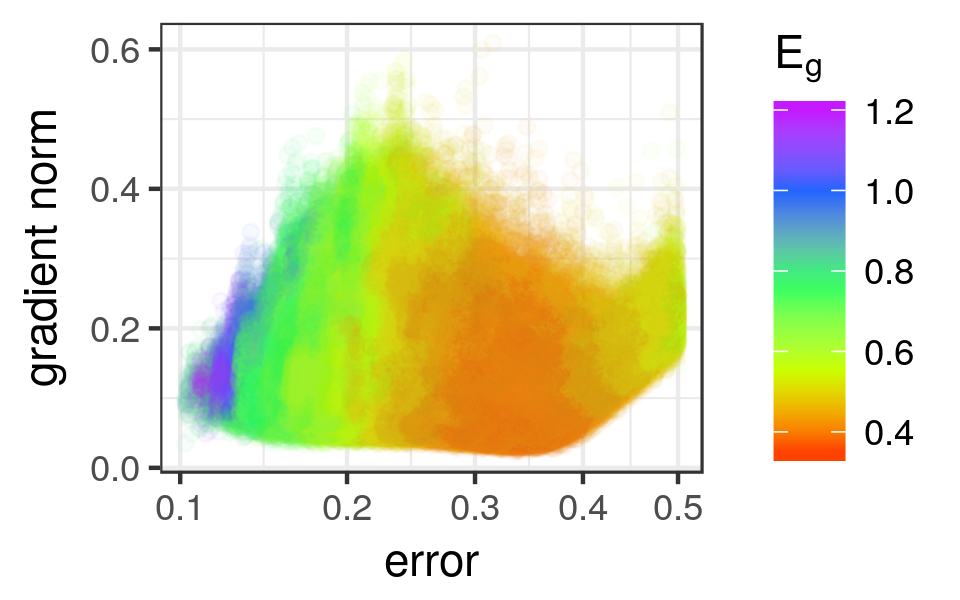}}
		\end{subfloat}
		\caption{LGCs colourised according to $E_g$ values for Heart, micro PGWs, $[-1,1]$ initialisation, $E_t< 0.05$.}\label{fig:heart:gen:act}
	\end{center}
\end{figure}

Fig.~\ref{fig:heart:gen:act} shows the LGCs colourised according to $E_g$ values. For the $[-1,1]$ walks, all activation functions yielded deteriorating $E_g$ values as $E_t$ approached zero. However, the band of good solutions was noticeably wider for ReLU and ELU as compared to TanH, indicating that it was easier to find a good quality solution on the ReLU and ELU loss landscapes. 

\subsection{MNIST}\label{sec:act:mnist}
\begin{figure}[!b]
	\begin{center}
		\begin{subfloat}[{TanH, micro }\label{fig:mnist:b1:tanh:micro}]{    \includegraphics[width=0.31\linewidth]{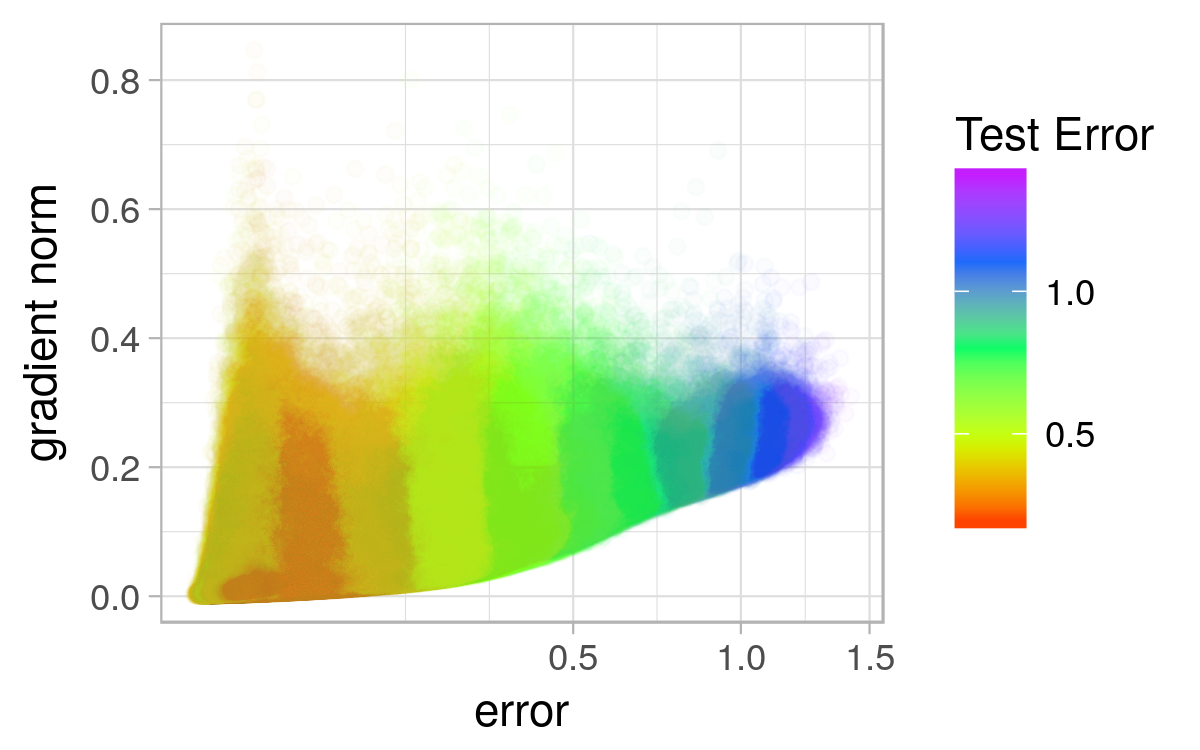}}
		\end{subfloat}
		\begin{subfloat}[{ReLU, micro }\label{fig:mnist:b1:relu:micro}]{    \includegraphics[width=0.31\linewidth]{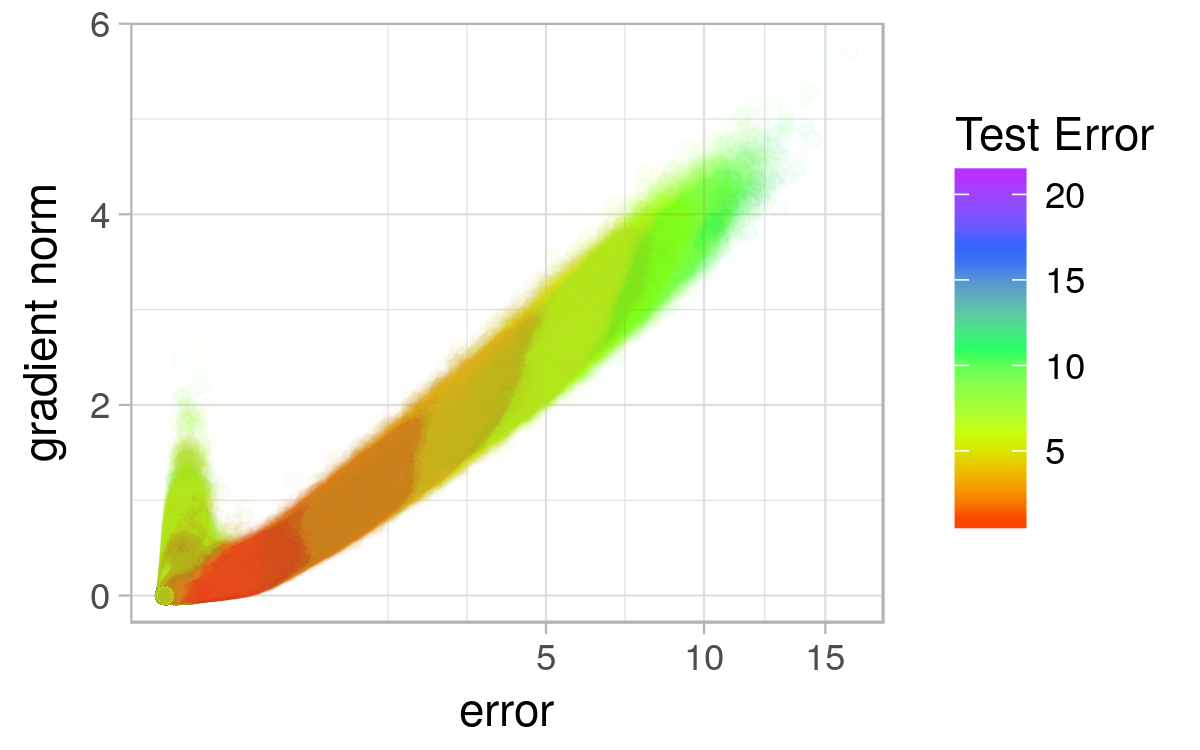}}
		\end{subfloat}		
		\begin{subfloat}[{ELU, micro }\label{fig:mnist:b1:elu:micro}]{    \includegraphics[width=0.31\linewidth]{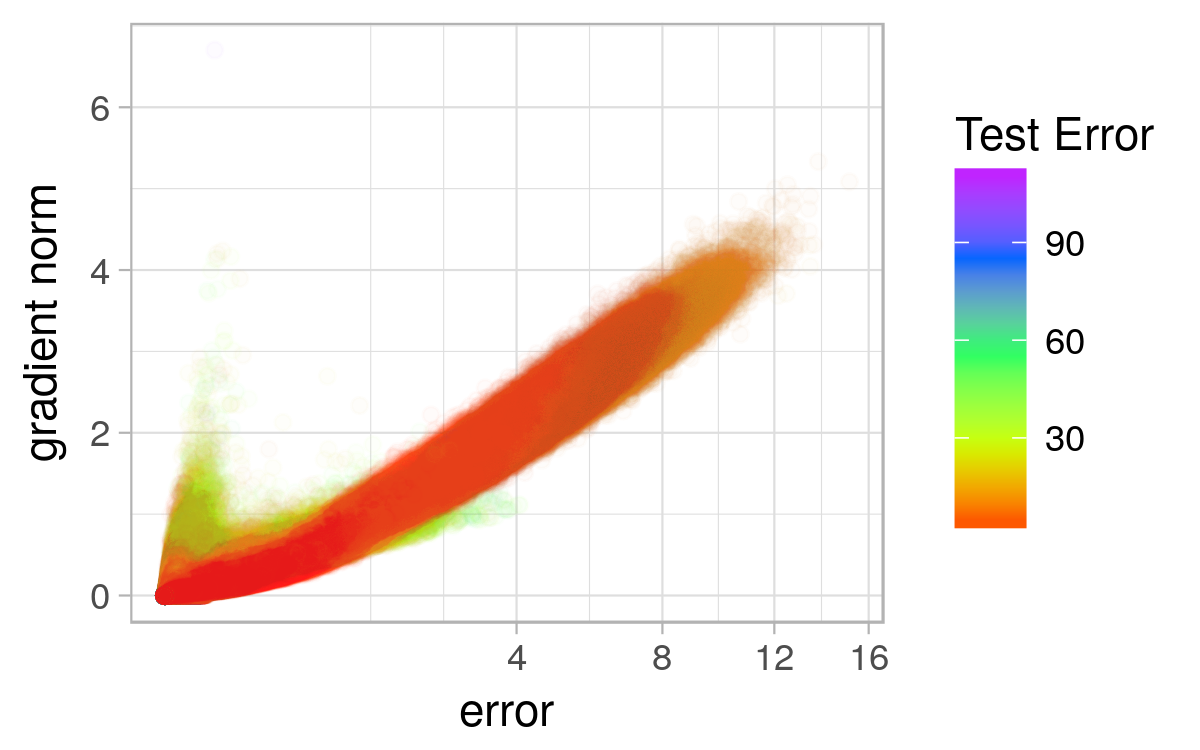}}
		\end{subfloat}			
		\begin{subfloat}[{TanH, macro }\label{fig:mnist:b1:tanh:macro}]{    \includegraphics[width=0.31\linewidth]{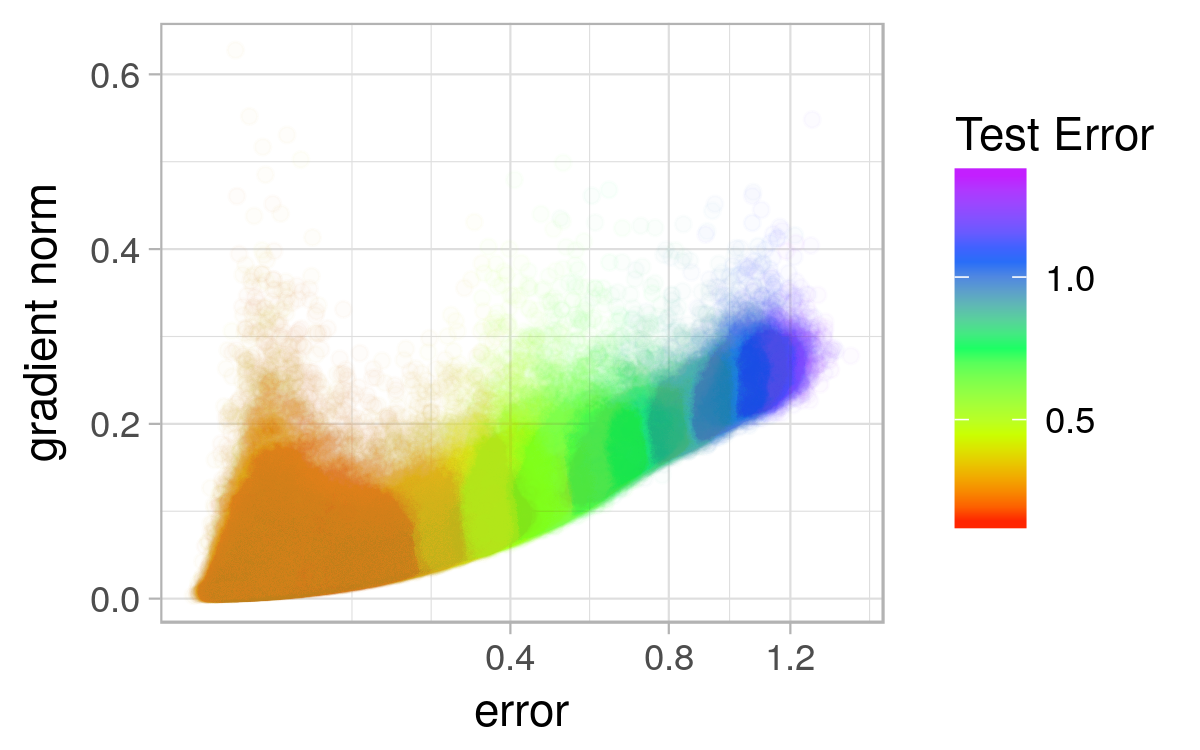}}
		\end{subfloat}	
		\begin{subfloat}[{ReLU, macro }\label{fig:mnist:b1:relu:macro}]{    \includegraphics[width=0.31\linewidth]{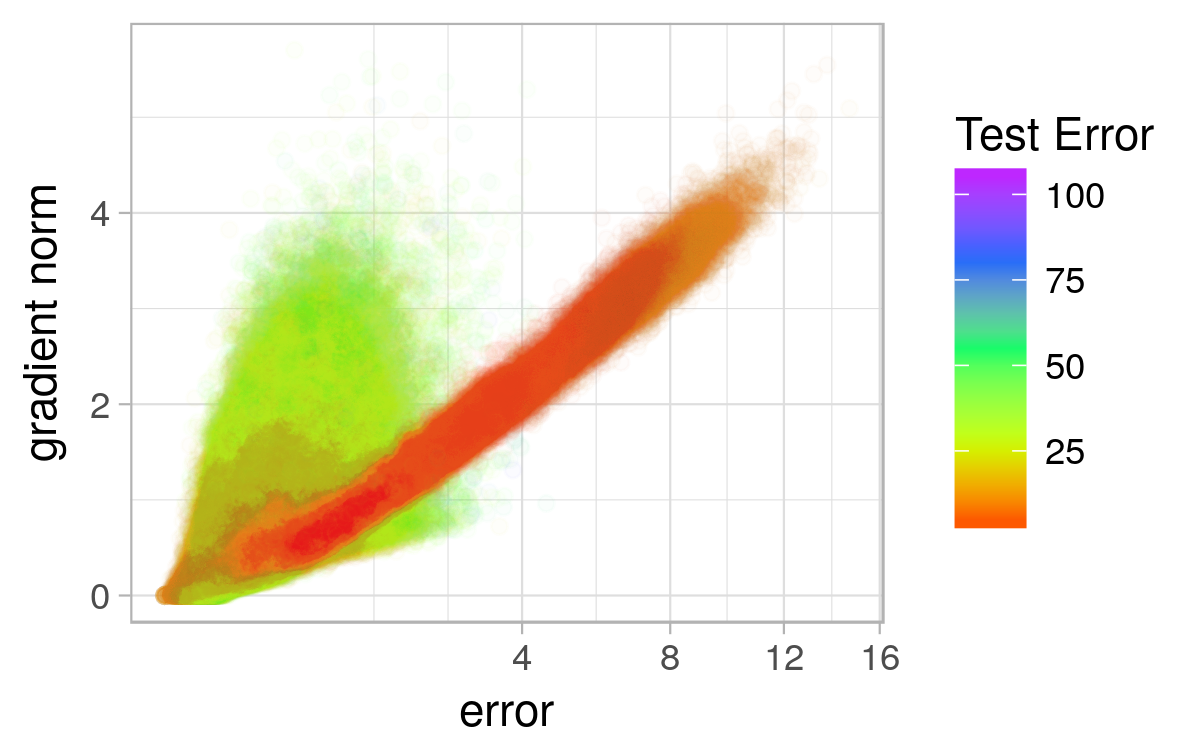}}
		\end{subfloat}
		\begin{subfloat}[{ELU, macro }\label{fig:mnist:b1:elu:macro}]{    	\includegraphics[width=0.31\linewidth]{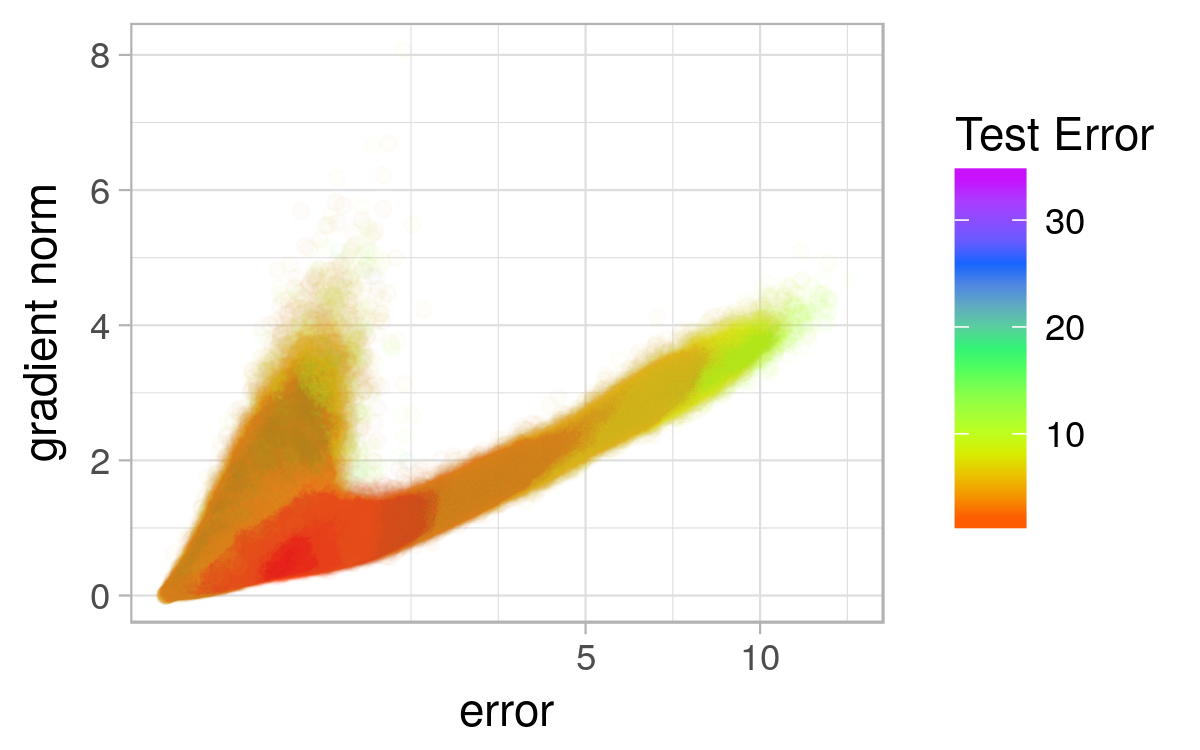}}
		\end{subfloat}
		\caption{LGCs for PGWs initialised in the $[-1,1]$ range for MNIST.}\label{fig:mnist:b1:micmac:act}
	\end{center}
\end{figure}

\begin{figure}[!b]
	\begin{center}
		\begin{subfloat}[{TanH, micro }\label{fig:mnist:b10:tanh:micro}]{    \includegraphics[width=0.31\linewidth]{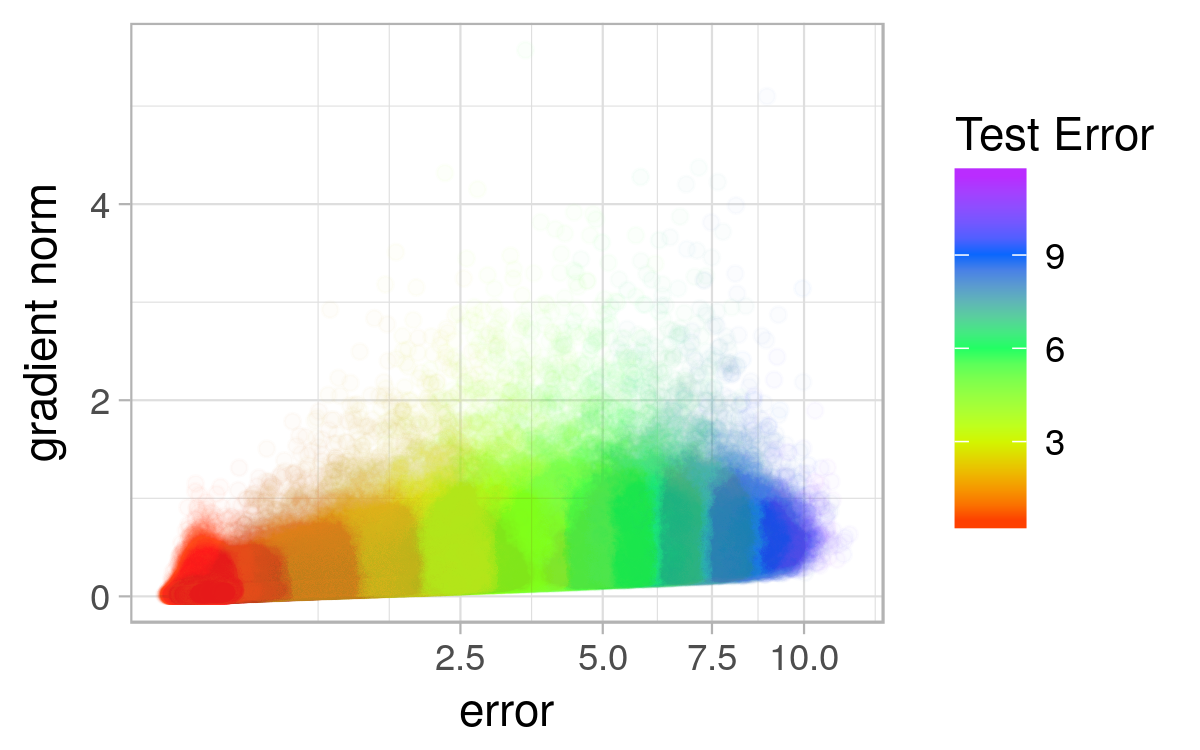}}
		\end{subfloat}
		\begin{subfloat}[{ReLU, micro }\label{fig:mnist:b10:relu:micro}]{    \includegraphics[width=0.31\linewidth]{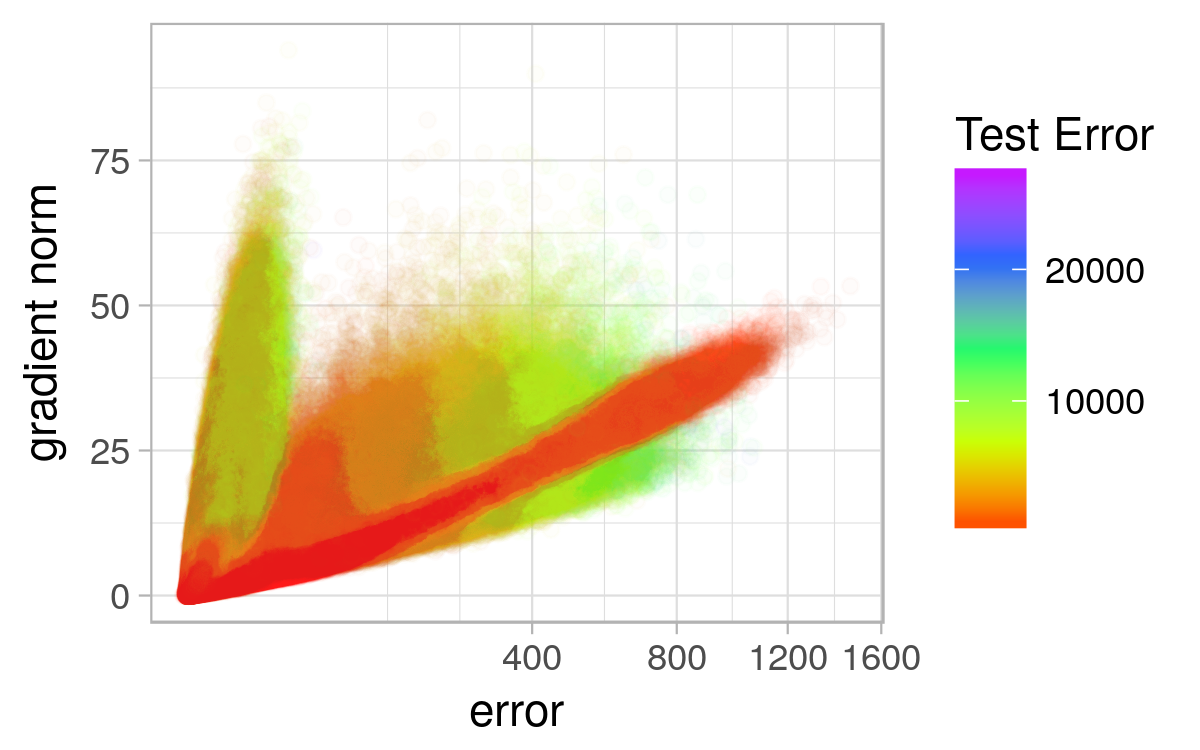}}
		\end{subfloat}	
		\begin{subfloat}[{ELU, micro }\label{fig:mnist:b10:elu:micro}]{    \includegraphics[width=0.31\linewidth]{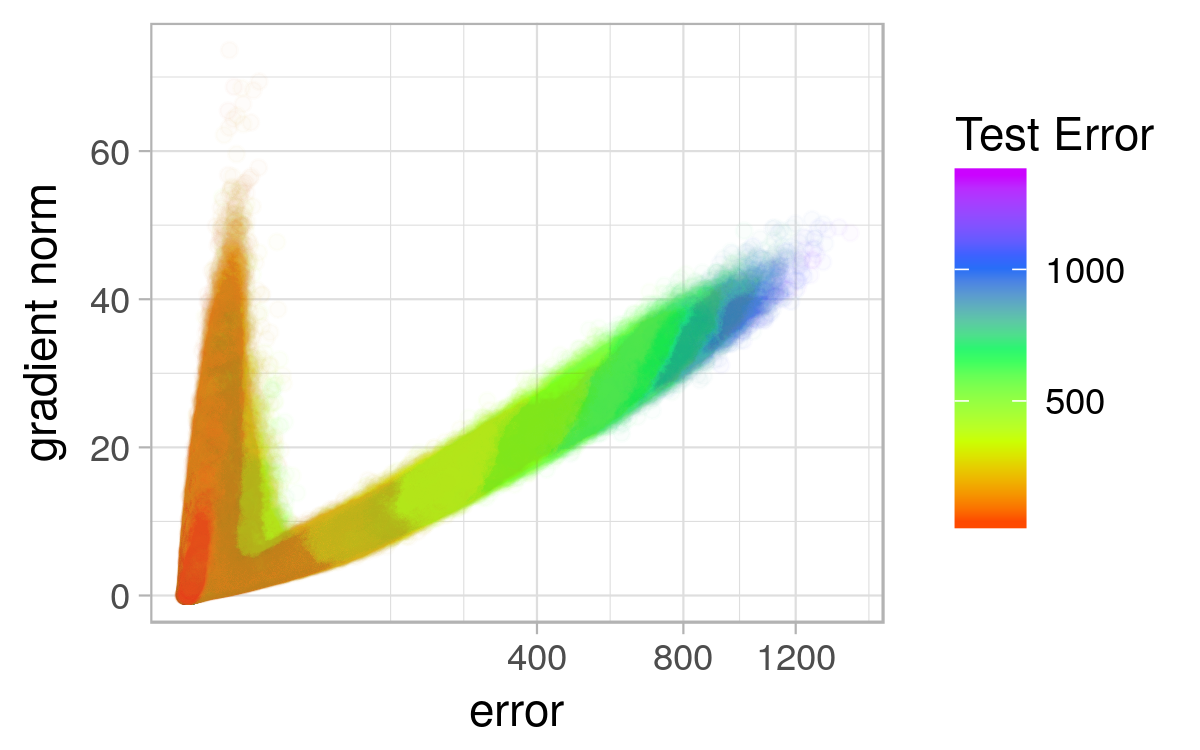}}
		\end{subfloat}
		\begin{subfloat}[{TanH, macro }\label{fig:mnist:b10:tanh:macro}]{    \includegraphics[width=0.31\linewidth]{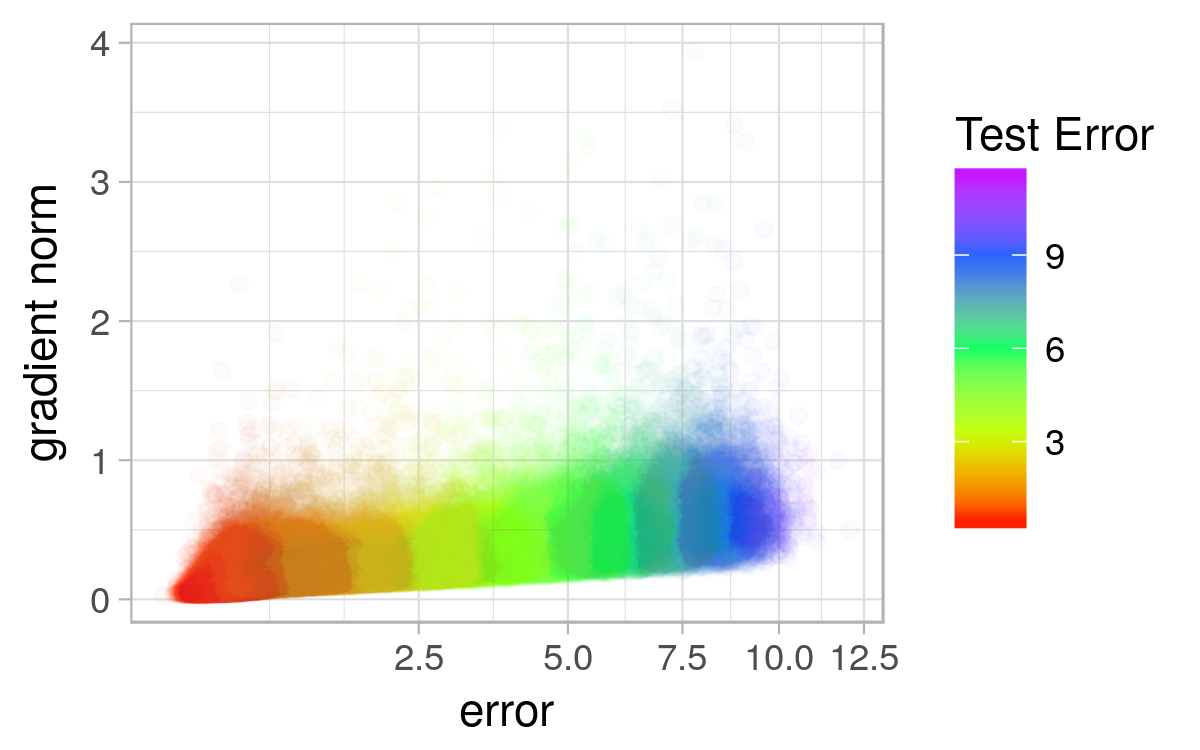}}
		\end{subfloat}
		\begin{subfloat}[{ReLU, macro }\label{fig:mnist:b10:relu:macro}]{    \includegraphics[width=0.31\linewidth]{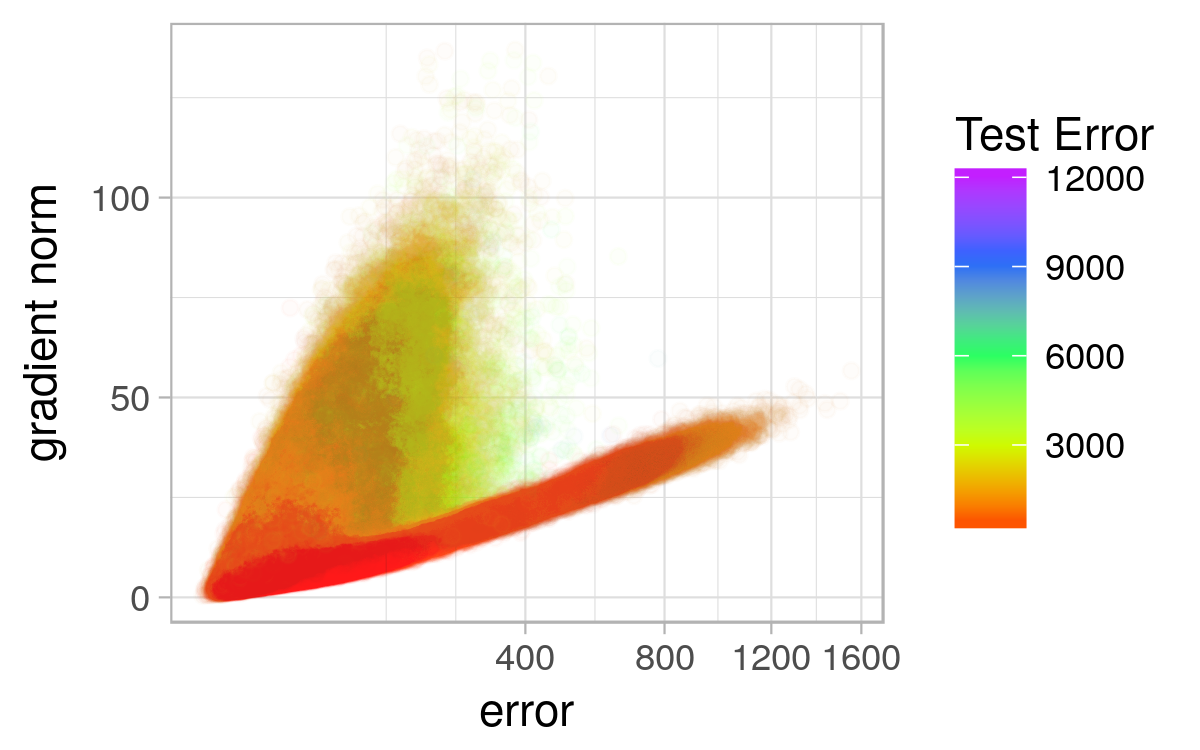}}
		\end{subfloat}
		\begin{subfloat}[{ELU, macro }\label{fig:mnist:b10:elu:macro}]{    \includegraphics[width=0.31\linewidth]{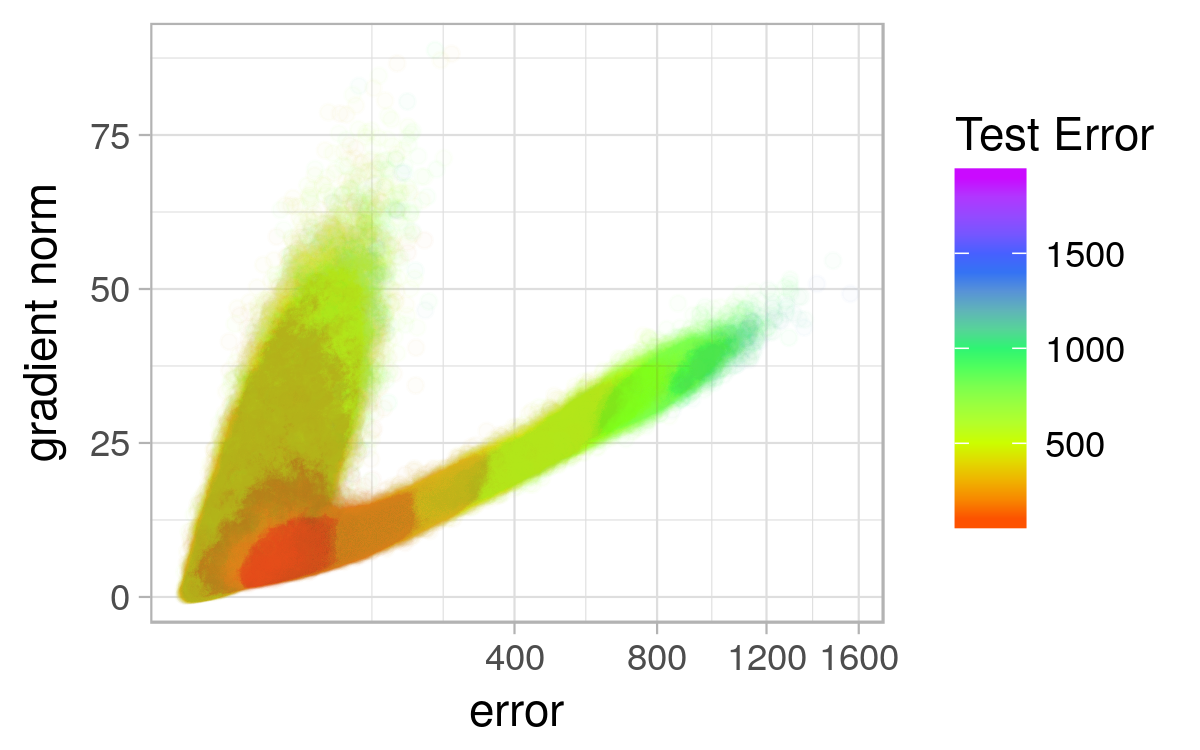}}
		\end{subfloat}
		\caption{LGCs for PGWs initialised in the $[-10,10]$ range for MNIST.}\label{fig:mnist:b10:micmac:act}
	\end{center}
\end{figure}

Due to the high dimensionality of MNIST, Hessians were not calculated for this experiment. 
Fig.~\ref{fig:mnist:b1:micmac:act} and \ref{fig:mnist:b10:micmac:act} show the LGCs colourised according to $E_g$ values. Convergence to a single global attractor is evident for all activation functions. 
Out of the activation functions considered, ELU once again yielded the most consistent generalisation performance.

Fig.~\ref{fig:mnist:b1:micmac:act} shows for the $[-1,1]$ interval that ReLU and ELU yielded much higher gradients than TanH. The two-cluster split was evident for all activation functions. For ReLU and ELU, the steep gradient cluster generally yielded poorer generalisation performance than the shallow gradient cluster. This behaviour is especially evident for ReLU under the macro setting (see Fig.~\ref{fig:mnist:b1:relu:macro}). The poor generalisation performance of high gradient points is attributed to neuron saturation.

Fig.~\ref{fig:mnist:b10:micmac:act} shows for the $[-10,10]$ range that TanH yielded an LGC without a well-formed structure. Low gradients and a noisy LGC indicate that the TanH error landscape was not very searchable under the $[-10,10]$ setting. 
ReLU and ELU also deteriorated with the increased step size and initialisation range, but not as drastically as TanH. The relative resilience of ReLU and ELU to the step size is attributed to the high gradients generated by these unbounded activation functions. ReLU and ELU evidently yielded a more searchable landscape for the high-dimensional MNIST problem, which correlates with the current deep learning insights~\cite{ref:Glorot:2010}. 

The split into two clusters became more pronounced for ReLU and ELU under the $[-10,10]$ setting, confirming the presence of narrow and wide valleys in the landscape. For ELU, points of good $E_g$ values were found in the steep cluster, likely due to the embedded minima discovered. ReLU yielded higher $E_t$ and $E_g$ values than ELU, confirming that the ELU error landscape was more resilient to overfitting.

\section{Conclusions}\label{sec:act:conclusions}
This paper empirically analysed the effect of using three different activation functions on the resulting NN error landscapes: TanH, ReLU, and ELU, used in the hidden layer. All experiments were conducted under four granularity settings, with different step sizes and initialisation ranges.

The choice of activation function did not have an effect on the total number of unique attractors (local minima) in the search space, but affected the properties of the discovered basins of attraction. ReLU and ELU yielded steeper attraction basins with stronger gradients than TanH. ReLU exhibited the most convexity, and ELU exhibited the least flatness. The stationary points exhibited by ReLU and ELU were generally more connected than the ones exhibited by TanH, indicating that ReLU and ELU yield more searchable landscapes. However, ReLU and ELU exhibited stronger sensitivity to the step size and the initialisation range than TanH. 

All activation functions yielded a split into high error, low gradient, and high gradient, low error clusters. The high gradient cluster was associated with indefinite (i.e. flat) curvature, caused by inactive, or non-contributing weights. Sample points in this cluster were often associated with poor generalisation. Thus, high (or steep) gradients were attributed to narrow valleys, associated with saturated neurons. In individual cases, points of high generalisation performance were discovered in the steep gradient clusters. These were attributed to the embedded self-regularised minima.

Out of the three activation functions considered, ELU consistently exhibited superior generalisation performance. Thus, the loss landscape yielded by the ELU activation function was the most resilient to overfitting.

In future, a more in-depth study of the self-regularised solutions will be conducted. If their unique loss landscape properties can be determined, algorithms may be developed that converge to areas with good generalisation properties that contain smaller NN architectures.

\begin{acks}
This research was supported by the National Research Foundation (South Africa) Thuthuka Grant Number 13819413. The authors acknowledge the Centre for High Performance Computing (CHPC),
South Africa, for providing computational resources to this research project.

\end{acks}

\bibliographystyle{ACM-Reference-Format}
\bibliography{acmart}

\end{document}